\documentclass[11pt]{article}
\usepackage[margin=1in]{geometry}
\usepackage{amsmath,amssymb}
\usepackage[T1]{fontenc}
\usepackage{lmodern}
\usepackage{microtype}
\usepackage{graphicx}
\usepackage[colorlinks=true,
            linkcolor=blue,
            citecolor=blue,
            urlcolor=blue]{hyperref}
\usepackage{booktabs}
\usepackage{array}

\providecommand{\doi}[1]{}
\renewcommand{\doi}[1]{}
\usepackage[numbers,square,sort&compress]{natbib}

\title{Bounded-Rationality, Hedging, and Generalization}
\author{Pedro A. Ortega\thanks{Daios Technologies, \texttt{pedro@daios.ai}}}
\date{}

\begin{document}

\maketitle

\begin{abstract}
A learner does not only fit data; it also determines how strongly 
the training sample may shape its output and how much distortion 
it can hedge. We study this relation as a bounded-rational decision
problem whose primitive object is the induced channel from samples
to outputs. The learner's response law determines which changes 
in this channel are cheap or costly, and therefore induces both
a lower tradeoff curve between training loss and sample dependence
and a matched upper certificate curve. When the response law is
represented by an $f$-divergence regularizer, these curves live
in the regularizer's native information geometry, with KL as
the special case corresponding to Shannon mutual information.
We show how the hedge and the two curves can be recovered from
black-box behavior by observing responses to scaled losses and
local loss perturbations. In learning, population loss is
empirical loss plus the distortion induced by the particular
training sample. The recovered hedge gives a practical
certificate when it covers that distortion. Thus generalization
is treated as a testable hedging property of the learner's
own response law.
\end{abstract}

\section{Introduction}
\label{sec:intro}

Overfitting can be viewed as exposure to distortion induced by the training sample.
Let $S$ denote the training sample and let $A$ denote the learner's
output. Fitting the sample allows~$S$ to influence~$A$. Stronger
dependence can improve empirical fit, but it also increases exposure to
structure that is specific to the sample and need not persist at
population level. Our question is whether the learner's own response
law controls that exposure.

Existing analyses of generalization study the same channel $P(A|S)$.
PAC-Bayesian theory bounds the generalization gap through a divergence
between the induced output distribution and an analyst-supplied prior
\citep{McAllester1999,Catoni2007}.
Stability analyses control it through the sensitivity of the output to
individual training points.
Information-theoretic approaches bound it in terms of the mutual
information $I(S;A)$ between sample and output
\citep{XuRaginsky2017,SteinkeZakynthinou2020,HellstromDurisiGuedjRaginsky2024}.
Each approach asks how large the generalization gap can be, given
the channel.
What the channel itself implies about that gap, absent a separately
supplied reference, remains a different question.

We identify the learner's cost structure with its native (or implicit) regularizer 
using the bounded-rational view of decision making.
In bounded rationality, an observation does not simply select the loss-minimizing
action; it changes the action law away from its marginal distribution, 
and this deviation carries a cost
\citep{OrtegaBraun2013,OrtegaBraunDyerKimTishby2015}. The operating
parameter $\beta$ sets the exchange rate between loss and that cost:
larger $\beta$ makes sample dependence cheaper, while smaller $\beta$
makes it more costly. For a fixed regularizer, changing $\beta$ selects
different operating points on the same tradeoff curve. Crucially,
no value of $\beta$ is singled out: different
values correspond to different hedging attitudes. In this paper we
study the case in which the deviation cost is represented by an 
$f$-divergence regularizer
\citep{Morimoto1963,AliSilvey1966,Csiszar1967,Vajda1968}. KL is the
special case whose native information coordinate is Shannon mutual
information \citep{Shannon1959,Berger1971}.

\begin{figure}[!tb]
    \centering
    \includegraphics[width=\linewidth]{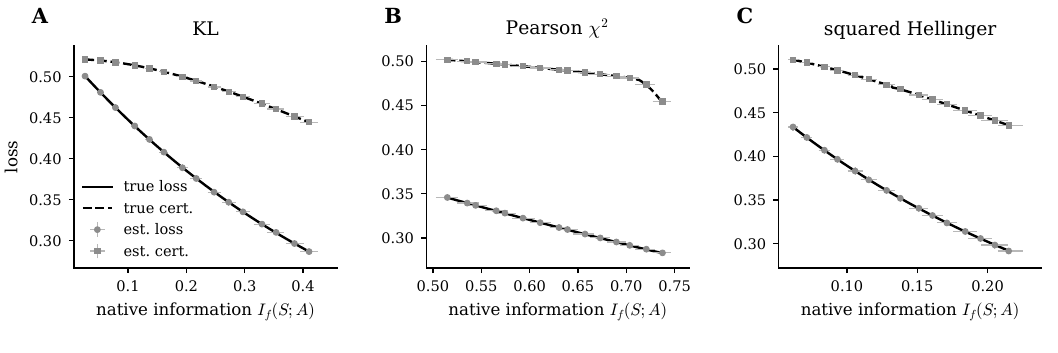}
    \caption{
    \textbf{Recovering a learner's native frontier and hedge from behavior.}
    Three learners solve the same categorical task, each governed by a
    different innate regularizer:
    \textbf{A}: KL,
    \textbf{B}: Pearson $\chi^2$, and
    \textbf{C}: squared Hellinger.
    For each learner, the solid black curve is the loss frontier: the best
    attainable loss at each level of native information use. The dashed
    black curve is the certificate frontier: the protected loss level
    implied by the learner's own hedge. Gray circles and squares show the
    same two curves recovered from black-box samples using the interventions
    in Section~\ref{sec:estimation}. Error bars show bootstrap
    $10$--$90$ percent intervals. The agreement shows the central claim of
    the paper: the response law encodes the regularizer, and therefore
    allows the native information coordinate, the loss frontier, and the
    hedging certificate to be recovered from behavior.
    }
    \label{fig:estimated-certificate}
\end{figure}

The same regularizer generates a \emph{native information coordinate} and two
paired curves: the \emph{lower frontier} and the \emph{certificate frontier}
(Figure~\ref{fig:estimated-certificate}). The lower frontier gives the
smallest attainable loss at a given level of native information use, and
comes from the bounded-rational tradeoff between loss and information.
The certificate frontier gives the loss level protected by the hedge
induced by the same regularizer, and comes from the dual interpretation
of information cost as an adversarial perturbation of the loss
\citep{Shannon1959,Berger1971,BenTalTeboulle1986,OrtegaLee2014}.
Taken together, these curves characterize the decision maker's implicit
hedging attitude: how much attainable loss it is willing to trade for
protection against regularizer-native perturbations.
The same construction applies to learning because learning can be
viewed as a bounded-rational decision problem. In this view, the
gap between empirical and population loss is an adversarial
perturbation of the learner's objective. The hedge identifies
the adversarial class induced by the learner; population loss is
controlled when the perturbation is covered by that class.

Because the frontiers are implicit properties of the learner's response
law, they can be reverse engineered from behavior through controlled
experiments, as shown in Figure~\ref{fig:estimated-certificate}.
Varying the task and operating conditions lets us estimate the native
operating curve and its matched certificate curve on the range explored
by the learner's responses, similarly to
\citep{OrtegaStocker2016,CaplinDean2015}. Since different regularizers
induce different information geometries, these curves are recovered
first in the native coordinate system selected by the response law. For
cross-regularizer comparison, the same recovered channels can then be
projected into a common information--loss plane, such as the Shannon
mutual-information plane.

The paper is organized as follows.
Section~\ref{sec:bounded-rationality} formulates bounded-rational
acting with a general regularizer and its native information coordinate.
Section~\ref{sec:adversarial-duality} uses duality to obtain the
endogenous hedge and the matched certificate frontier.
Section~\ref{sec:comparing-dms} describes the resulting lower and
certificate frontiers and explains how decision makers can be compared
by projecting their channels into a common information--loss
plane.
Section~\ref{sec:estimation} shows how the lower and
certificate curves can be recovered from black-box behavior.
Section~\ref{sec:generalization} then applies the same construction to
learning.

\section{Bounded Rationality}
\label{sec:bounded-rationality}

We consider a decision problem in which a stimulus random variable $S$
is drawn from a distribution~$P(S)$ and observed by a decision maker,
who then produces an action random variable~$A$ from~$P(A |S)$.
For realizations $S=s$ and $A=a$, the incurred loss is
$\ell(s,a)$. The bounded-rational premise is that the decision maker
does not choose the channel $P(A|S)$ by minimizing loss alone. Instead,
it trades off loss against the cost of allowing the stimulus to reshape
the action distribution. More precisely, in information-theoretic bounded rationality the objective is the free energy functional
\begin{equation}
    \label{eq:avg-free-energy}
    F_\beta[P(A|S)] :=
    \int P(s) \biggl[ 
        \underbrace{
        \int P(a|s)\,\ell(s,a)\,da
        }_\text{expected loss}
        + \underbrace{
        \frac{1}{\beta}\, D\bigl(P(a|s)\bigl\| P(a)\bigr)
        }_\text{regularization}
    \biggr]\, ds,
\end{equation}
where
\begin{equation}
    \label{eq:marginal-self-consistency}
    P(a) = \int P(a|s)\,P(s)\,ds.
\end{equation}
Here~$\beta>0$ converts information cost into loss units and  
$D(Q\|P)$ is the KL-divergence. Write the expected loss as
\begin{equation}
    \label{eq:expected-loss}
    L := \int\!\!\int P(s,a)\,\ell(s,a)\,da\,ds.
\end{equation}
Thus each choice of the operating level $\beta$
yields a different tradeoff between expected loss
and the informational cost of conditioning on the stimulus,
with no privileged choice of~$\beta$ \citep{OrtegaBraun2013,BraunOrtega2014,OrtegaBraunDyerKimTishby2015}.
The resulting optimum is the \emph{certainty-equivalent value} of
the choice, i.e.\ the single loss value that summarizes the value of
the whole bounded-rational decision problem.

We now generalize the regularization to $f$-divergences~\citep{Morimoto1963,AliSilvey1966,Csiszar1967,Vajda1968}. 
This choice is for mathematical convenience, although we expect
the results to generalize to a broader class. Fix a convex lower-semicontinuous
function $f:[0,\infty)\to(-\infty,\infty]$ with $f(1)=0$. If $Q(A)$ and $P(A)$ are
distributions on the action space, define the $f$-divergence
\begin{equation}
    \label{eq:f-div}
    D_f(Q\|P) := \int P(a)\,f\!\left(\frac{Q(a)}{P(a)}\right)\,da,
\end{equation}
with the usual convention that the value is $+\infty$ whenever the
ratio $Q(a)/P(a)$ is not admissible where $Q(a)$ is positive.
This class preserves convexity and dual structure while allowing the
geometry of the information penalty to vary across models. 
Statements in the following sections
that use only Fenchel duality extend further to general proper convex
regularizers \citep{Rockafellar1970,BenTalTeboulle1987,LieseVajda2006}.

We furthermore fix canonical representatives for the regularizers. 
Each $f$-divergence has an equivalence class via $f(x) \mapsto f(x) + c(x-1)$
that yields the same divergence \citep{AliSilvey1966,Csiszar1967,LieseVajda2006}.
Therefore, whenever $f$ is differentiable
at $1$, we choose the representative with $f'(1)=0$, 
which fixes an additive gauge while preserving
the divergence geometry. With this convention, the KL divergence
is obtained from
$f(x)=x\log x-x+1$. Other examples are the Pearson $\chi^2$ 
divergence, obtained with $f(x)=(x-1)^2$, and the squared 
Hellinger divergence, obtained with $f(x)=(\sqrt{x}-1)^2$.
In what follows, the regularizer is restricted to this class. 

\begin{figure}[!tb]
\centering
\includegraphics[width=\textwidth]{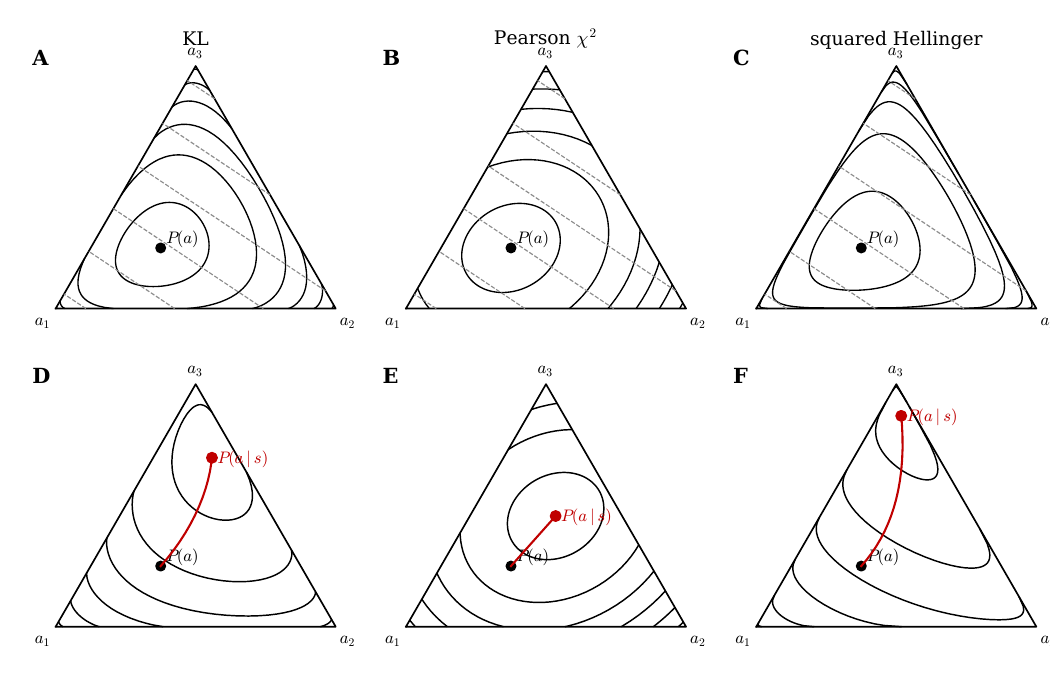}
\caption{\textbf{Geometry of bounded-rational acting under three regularizers.}
\textbf{A--C:} probability simplex over three actions, showing the prior action distribution $P(a)$ together with the loss and regularizer geometries for KL, Pearson $\chi^2$, and squared Hellinger. Gray dashed lines are level sets of the loss; black solid curves are level sets of the corresponding divergence.
\textbf{D--F:} the same three simplices after combining the loss with the regularizer at an operating level $\beta$.
Black solid curves are level sets of the bounded-rational objective.
The black point marks the prior $P(a)$, the red point marks the optimizer $P(a\mid s)$, and the red curve is the bounded-rational operating path traced by varying $\beta$ from $0$ to that value.
The loss is the same in all panels, but the regularizer changes both the local geometry around $P(a)$ and the displacement toward the bounded-rational optimum.}
\label{fig:geometry}
\end{figure}

\begin{figure}[!tb]
\centering
\includegraphics[width=\textwidth]{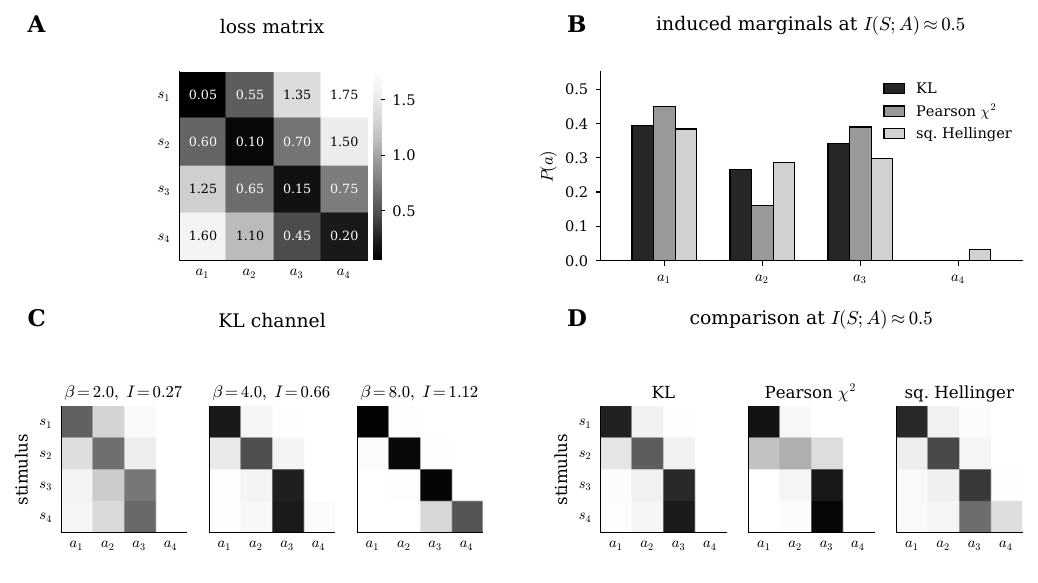}
\caption{\textbf{Bounded-rational acting in a finite categorical model.}
\textbf{A:} A categorical choice task defined by a $4\times 4$ 
loss matrix $\ell(s,a)$.
\textbf{B:} The induced marginal action distribution $P(a)$ for KL,
Pearson $\chi^2$, and squared Hellinger regularization at a matched
information level, $I(S;A)\approx 0.5$.
\textbf{C:} The KL channel $P(a\mid s)$ at various operating points;
increasing the operating level concentrates mass more strongly on
low-loss actions.
\textbf{D:} Comparison of the channels induced by KL, Pearson~$\chi^2$, 
and squared Hellinger on the same task at matched information.
The decision maker does not collapse to the pointwise loss minimizer.
Instead it randomizes over actions, and the form of that randomization depends on the regularizer.}
\label{fig:hedging}
\end{figure}

Averaging the $f$-divergence over the stimulus produces a corresponding
native mutual information. Indeed, define the \emph{$f$-mutual information}
between $S$ and $A$ as
\begin{equation}
    \label{eq:f-info}
    I_f(S;A)
    :=
    D_f\bigl(P(S,A)\|P(S)P(A)\bigr)
    =
    \int P(s)\,
    D_f\bigl(P(A|s)\|P(A)\bigr)\,ds,
\end{equation}
where $P(s,a)=P(s)P(a|s)$ \citep{Csiszar1972,ZivZakai1973,Csiszar1995,SasonVerdu2018}.
This is the native information coordinate
selected by the regularizer $f$. Only in the KL case does this native
information coordinate reduce to Shannon mutual information \citep{Shannon1959,Berger1971}.
In this sense, the choice of regularizer determines what counts as dependence
between stimulus and action, and therefore fixes the native geometry of fit (Figure~\ref{fig:geometry}), and, as we will see in the next section, hedging (Figure~\ref{fig:hedging}). Appendix~\ref{app:additional-f-divergences} lists additional regularizers.

\section{Hedging}
\label{sec:adversarial-duality}

The bounded-rational acting problem admits an equivalent adversarial
interpretation \citep{OrtegaLee2014}. Through Fenchel duality, the objective
can be restated as a minimax game against an endogenous adversary.
More specifically, the regularizer encodes a hedging strategy: it fixes
the geometry of admissible adversarial perturbations, while $\beta$ sets their
effective scale. Smaller values of $\beta$ correspond to larger
perturbations and therefore to stronger hedging.

We will develop this idea in two steps. First, fix a stimulus $S=s$ and 
a prior~$P(A)$. By
Fenchel duality \citep{Fenchel1949,Rockafellar1970,BenTalTeboulle1987},
the regularizer can be expressed as
\begin{equation}
    \label{eq:div-dual}
    D_f\bigl(P(A|s)\|P(A)\bigr)
    =
    \sup_C
    \left\{
        \beta\int P(a|s)\,C(a)\,da
        - \int P(a)\,f^\star(\beta C(a))\,da
    \right\},
\end{equation}
where the supremum is over cost functions $C$ for which the integrals
are well defined, and $f^\star$ is the convex conjugate of $f$.
Substituting this into the free-energy functional gives
\begin{equation}
    \label{eq:fe-dual}
    F_\beta[P(A|S=s)]
    =
    \sup_C
    \biggl\{
        \underbrace{
            \int P(a|s)\,\bigl(\ell(s,a)+C(a)\bigr)\,da
        }_\text{perturbed loss}
        - \underbrace{
            \frac{1}{\beta}\int P(a)\,f^\star(\beta C(a))\,da
        }_\text{adversarial penalty}
    \biggr\}.
\end{equation}
This reveals that the information penalty is equivalent to an adversary
that adds the perturbation $C$ to the nominal loss, but must pay a
penalty determined by $f^\star$ under the marginal action distribution~$P(A)$
\citep{OrtegaLee2014,BenTalTeboulle1987}. Notice that for a fixed
perturbation~$C$, the adversarial penalty
\[
    \frac{1}{\beta}\int P(a)\,f^\star(\beta C(a))\,da
\]
is constant with respect to the decision maker's choice of
$P(A|s)$. Thus the decision maker's best response to a given
perturbation $C$ is determined entirely by the \emph{effective loss}
\begin{equation}
    \label{eq:effective-loss}
    \ell(s,a)+C(a).
\end{equation}

Our next step generalizes this idea to the channel level. At this level,
changing $P(A|S)$ also changes the marginal action distribution $P(A)$.
Thus the adversarial perturbation must account not only for the direct
change in the conditional law $P(a|s)$, but also for the induced change
in the shared marginal $P(a)$.

The guiding idea is the same as before. A perturbation of the loss
changes the objective through its expected value under the channel,
whereas a perturbation of the channel changes the native information
penalty. The adversarial perturbation is therefore the loss perturbation
whose first-order effect matches the first-order effect of the native
information cost. In this sense, the adversary reflects the gradient of
the decision maker's own information use back into the loss. The 
factor~$1/\beta$ converts this information gradient into loss units: smaller
$\beta$ makes the same dependence more costly, and therefore corresponds
to a larger hedging perturbation.

Using calculus of variations under the self-consistency
constraint \eqref{eq:marginal-self-consistency}
gives the optimal channel-level perturbation. The derivation is given in
Appendix~\ref{app:optimal-perturbation}. Assuming that $f$ is
differentiable at every value $P(a|s)/P(a)$ appearing in the channel,
the adversary's optimal perturbation for stimulus~$s$ is
\begin{equation}
    \label{eq:opt-perturbation}
    C_s^{\mathrm{opt}}(a)
    =
    \frac{1}{\beta}
    \left[
        f'\!\left(\frac{P(a|s)}{P(a)}\right)
        +
        G(a)
    \right],
\end{equation}
where
\begin{equation}
    \label{eq:G-term}
    G(a)
    :=
    \int P(s)
    \left[
        f\!\left(\frac{P(a|s)}{P(a)}\right)
        -
        \frac{P(a|s)}{P(a)}
        f'\!\left(\frac{P(a|s)}{P(a)}\right)
    \right]ds .
\end{equation}
The first term in \eqref{eq:opt-perturbation} is the direct
likelihood-ratio contribution. The term $G(a)$ is the marginal
correction induced by the shared action distribution, and therefore
couples the perturbations across stimuli.

\begin{figure}[p]
    \centering

    \includegraphics[width=\textwidth]{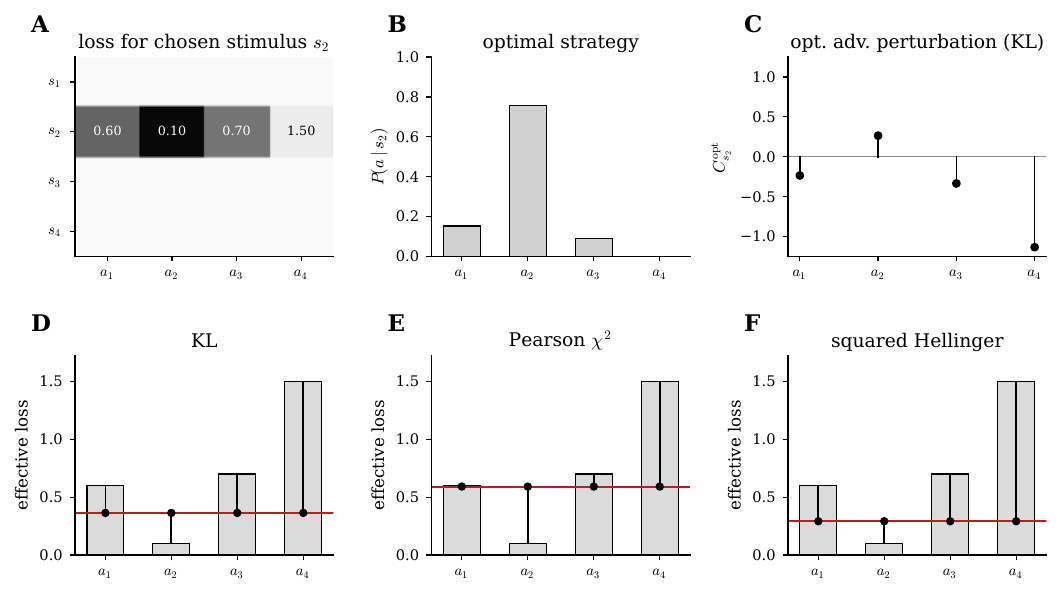}

    \caption{
    \textbf{Adversarial duality and indifference.}
    We continue with the problem from Figure~\ref{fig:hedging},
    and assume the chosen stimulus is $s_2$.
    \textbf{A:} the loss matrix, with the chosen row
    $\ell(s_2,a)$ highlighted.
    \textbf{B:} the KL bounded-rational strategy $P(a\mid s_2)$ at
    $\beta=4.0$.
    \textbf{C:} the corresponding optimal adversarial perturbation
    $C_{s_2}^{\mathrm{opt}}(a)$.
    \textbf{D--F:} effective-loss construction for KL, Pearson $\chi^2$, and
    squared Hellinger regularization.
    Gray bars show the nominal losses $\ell(s_2,a)$.
    Black impulses show the adversarial correction from nominal loss to the
    common indifference level (red line).
    }
    \label{fig:adversarial-duality}

    \vspace{1em}

    \includegraphics[width=\textwidth]{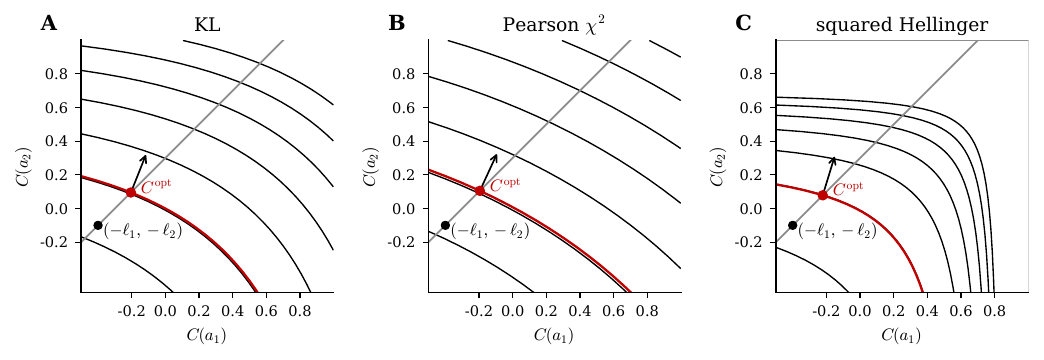}

    \caption{
    \textbf{Perturbation classes and geometric construction of policy.}
    \textbf{A--C:} KL, Pearson~$\chi^2$, and squared Hellinger induce different
    families of iso-penalty contours in the same two-action cost plane.
    The gray line is the indifference line
    $\ell_1 + C(a_1) = \ell_2 + C(a_2)$ determined by the nominal losses.
    The red point marks the optimal perturbation $C^{\mathrm{opt}}$, and the red
    curve is the iso-penalty contour passing through it, delimiting the
    adversarial class $\mathcal{C}$.
    The black arrow is the normal to that contour at $C^{\mathrm{opt}}$.
    Due to the indifference condition, this normal is the bounded-rational
    policy.
    }
    \label{fig:cost-penalty-geometry}

\end{figure}

\paragraph{Indifference.}
Bounded-rational choices obey an indifference condition
\citep{OrtegaLee2014}. Introducing a Lagrange multiplier $\lambda_s$
for the normalization constraint $\int P(a|s)\,da=1$, the
first-order condition for the channel objective on the support of
$P(a|s)$ is
\[
    \ell(s,a)
    +
    \frac{1}{\beta}
    \left[
        f'\!\left(\frac{P(a|s)}{P(a)}\right)
        +
        G(a)
    \right]
    =
    \lambda_s.
\]
Since $C_s^{\mathrm{opt}}(a)$ is~\eqref{eq:opt-perturbation}, this is exactly
\begin{equation}
    \label{eq:indifference}
    \begin{aligned}
    \ell(s,a)+C_s^{\mathrm{opt}}(a)=\lambda_s
    &\qquad\text{for every }a\text{ with }P(a|s)>0,\\
    \ell(s,a)+C_s^{\mathrm{opt}}(a)\ge\lambda_s
    &\qquad\text{for every }a\text{ with }P(a|s)=0,
    \end{aligned}
\end{equation}
that is, the effective losses~\eqref{eq:effective-loss} are equal,
where the inequality off the support follows from complementary
slackness. Figure~\ref{fig:cost-penalty-geometry} shows this construction in the two-action cost plane.

At equilibrium, the decision maker and the adversary are matched. For
each stimulus $s$, the decision maker assigns positive probability only
to actions whose effective loss is the
same. The adversary does this by adding more cost to actions with
smaller loss and less cost to actions with larger loss, subject to the
penalty imposed by the regularizer. Thus the decision maker is
indifferent among the actions it uses, and the adversary cannot increase
the common effective loss without paying a larger penalty. In finite
action spaces this gives the exact minimax reading of the hedge~\citep{Sion1958,Rockafellar1970}.

\paragraph{Certificates.}
The same duality yields a corresponding certificate curve
\citep{OrtegaLee2014,BrekelmansGeneweinGrauMoyaDeletangKuneschLeggOrtega2022}.
Every decision maker implicitly assumes a class of 
perturbations~$\mathcal{C}$, and hedges against it
by minimizing the effective loss~\eqref{eq:effective-loss}
under the worst-case perturbation~$C_s^{\mathrm{opt}}$ defined
in~\eqref{eq:opt-perturbation}. This class $\mathcal{C}$ is given by
\begin{equation}
    \label{eq:advset}
    \mathcal{C} 
    := \Bigl\{
        C_s \Bigm| 
        \Phi(C_s) \le \Phi(C^{\mathrm{opt}}_s)
    \Bigr\},
    \qquad\text{where}\qquad
    \Phi(C_s) := \frac{1}{\beta} \int\!\!\int P(s)P(a)\,f^\star\!\bigl(\beta C_s(a)\bigr)\,da\,ds,
\end{equation}
i.e.\ the loss perturbations covered by the endogenous hedge, as illustrated
in Figure~\ref{fig:cost-penalty-geometry}.
Accordingly, we define the certificate value
\begin{equation}
    \label{eq:certificate}
    \begin{aligned}
    L_f^{\mathrm{adv}}
    &:= \sup_{C_s \in \mathcal{C}} \int\!\!\int P(s,a) [\ell(s,a)+C_s(a)]\,da\,ds \\
    &= \int\!\!\int P(s,a) \left[\ell(s,a) + \frac{1}{\beta}
    \left\{
        f'\!\left(\frac{P(a|s)}{P(a)}\right) + G(a)
    \right\}\right]\,da\,ds.
    \end{aligned}
\end{equation}
This is the hedged loss induced by the chosen regularizer at operating
level $\beta$. The certificate is conditional on the perturbation lying
in the class $\mathcal{C}$ implied by the bounded-rational rule itself.
If the relevant perturbation is outside $\mathcal{C}$, the certificate
does not apply, and the loss could be arbitrarily bad.

To evaluate \eqref{eq:certificate}, it remains to compute $f'$ and
$G$. For the three regularizers introduced above, these terms are
\begin{equation}
\begin{aligned}
    \text{KL:}\qquad
    f'(x) &= \log x,
    & G(a) &= 0,
    \\[0.5em]
    \text{Pearson }\chi^2:\qquad
    f'(x) &= 2(x-1),
    & G(a) &= 1-\int P(s)
        \left(\frac{P(a|s)}{P(a)}\right)^2ds,
    \\[0.5em]
    \text{squared Hellinger:}\qquad
    f'(x) &= 1-\frac{1}{\sqrt{x}},
    & G(a) &= 1-\int P(s)
        \sqrt{\frac{P(a|s)}{P(a)}}\,ds.
\end{aligned}
\end{equation}
KL is special because $G(a)=0$ and the native information
coordinate is Shannon mutual information.

\begin{figure}[!tb]
    \centering
    \includegraphics[width=\linewidth]{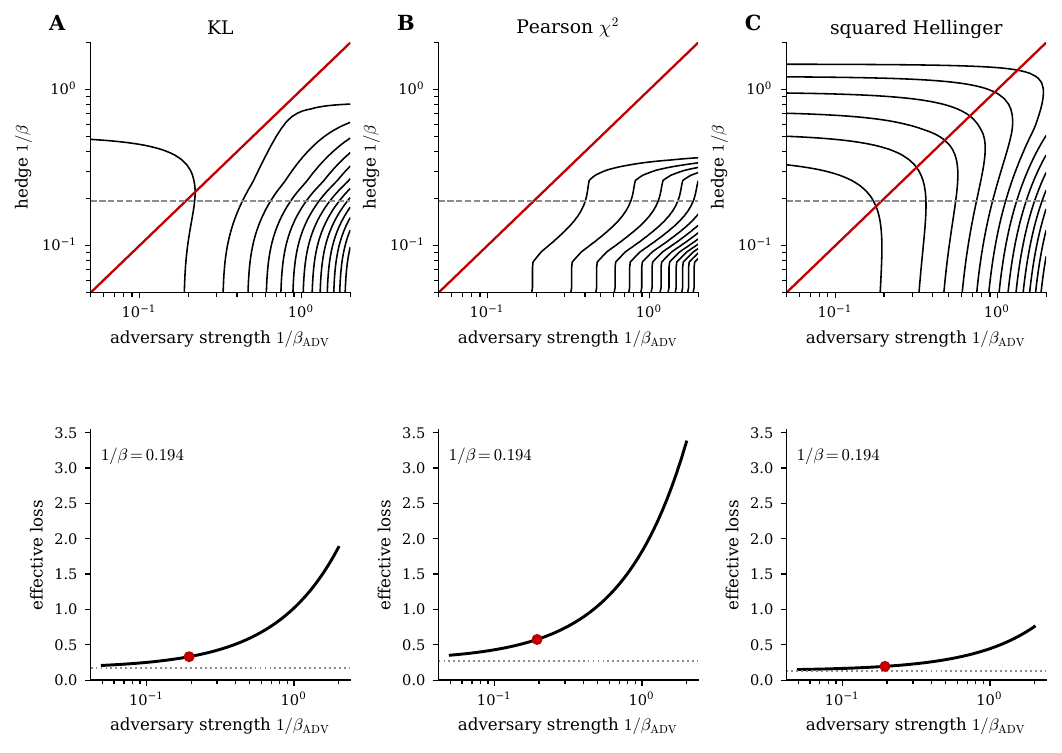}
    \caption{
    \textbf{Hedging against stronger perturbations.}
    The task from Figure~\ref{fig:hedging}
    is evaluated under KL, Pearson $\chi^2$, and squared Hellinger
    regularization. The contours in \textbf{A--C} show the effective
    loss obtained by evaluating the channel at hedge $1/\beta$
    against the strongest adversarial perturbation with
    $1/\beta_{\mathrm{ADV}}$. The red diagonal marks the perfect hedge 
    $\beta=\beta_{\mathrm{ADV}}$, which gives the certificate.
    Moving upward increases the decision-maker's hedge and 
    changes how rapidly the effective loss rises along horizontal slices.
    The effective loss grows particularly fast in the in the under-hedged
    region below the diagonal.
    Panels \textbf{D--F} show the effective loss along the gray horizontal
    slice in \textbf{A--C}. The red point is the certificate, and
    the gray dotted line marks the (unperturbed) loss.
    }
    \label{fig:hedging-contours}
\end{figure}

Finally, it can be shown that the average certificate equals
the native information cost:
\[
    L_f^{\mathrm{adv}}-L
    = \frac{1}{\beta}I_f(S;A).
\]
The derivation is given in Appendix~\ref{app:certificate-identity}.
Equivalently, the certificate $L_f^{\mathrm{adv}}$ coincides with the 
certainty-equivalent value $L+\tfrac{1}{\beta}I_f(S;A)$. 
Economically, this leads to a double reading of the gap: the certainty-equivalent
value computes this price as a regularization cost, while
the certificate computes the same price as the loss added by the
endogenous worst-case perturbation. Thus, the information cost
is the price of insurance against the exposure created by
conditioning on the stimulus.
Figure~\ref{fig:hedging-contours} illustrates the practical
meaning of the certificate. Each horizontal slice fixes the decision-maker hedge and 
varies the perturbation strength. Stronger decision-maker hedging changes 
these slices by trading loss for reduced exposure to perturbation.
The matched diagonal gives the certificate.

\paragraph{High-probability bounds.}
Although out of scope for our exposition, it is worth mentioning 
that the same duality also gives channel-wide
tail bounds for perturbations expressed in native information. Let
$C_s(a)$ be a perturbation, fix a threshold $u$, and define the tail
event
\[
    E=\{C_S(A)>u\}.
\]
Let $q$ and $p$ be the probability of the tail event under the product law
$P(S)P(A)$ and under the channel law $P(S,A)$:
\[
    q=\int_E P(s)P(a)\,da\,ds,
    \qquad p=\int_E P(s,a)\,da\,ds.
\]
The strategy follows two steps. First, we bound the tail event under
the product law: $q\le \bar q$. 
This can be done by the same dual transform associated with the
regularizer, or by any sharper problem-specific tail bound that gives $\bar{q}$. 
Second,
we transfer this product-law tail bound to the channel law. By the data
processing inequality for $f$-divergences applied to the event $E$, and by
the monotonicity of the resulting binary divergence in $q$ for $q\le p$,
any $p>\bar q$ must satisfy
\begin{equation}
    \label{eq:bindiv-mi}
    \bar q f\!\left(\frac{p}{\bar q}\right)
    + (1-\bar q)f\!\left(\frac{1-p}{1-\bar q}\right)
    \le I_f(S;A).
\end{equation}
The derivation is given in Appendix~\ref{app:tail-transfer}, along with
examples of concrete tail bounds. Consequently,
\[
    P(C_S(A)>u)=p\le\delta_f(\bar q),
\]
where $\delta_f(\bar q)$ is the largest $p\in[\bar q,1]$ satisfying the
last inequality~\eqref{eq:bindiv-mi}. Thus the product-law tail bound 
controls how often the perturbation is large without dependence, 
and the native information controls how much the channel can amplify
that event.

\section{Frontiers, and comparing decision makers}
\label{sec:comparing-dms}

The previous sections associated each channel with three
quantities: its native information use, its loss, and its 
certificate value. Varying the operating level for a fixed
regularizer therefore produces two paired curves in the
information--loss plane that characterize the performance trade--offs
of the decision maker: a lower curve of attainable losses and 
an upper curve of certified losses. This section defines 
these frontiers and explains how to
compare decision makers governed by different regularizers.

\begin{figure}[!tb]
    \centering
    \includegraphics[width=\linewidth]{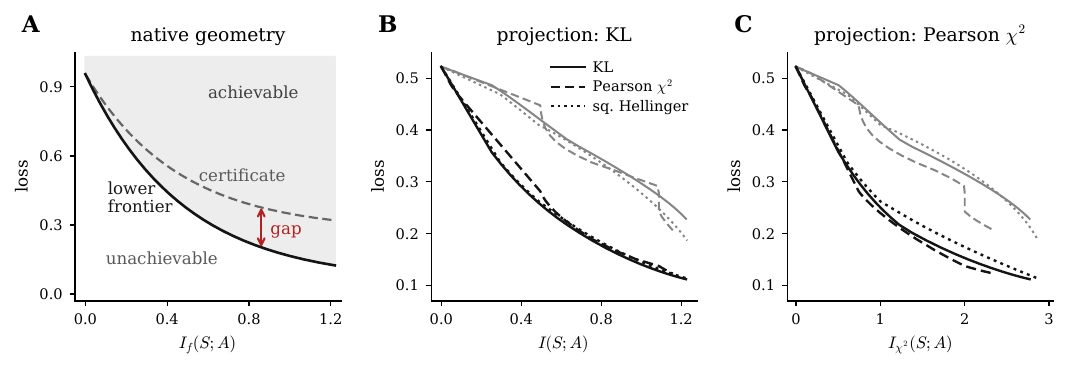}
    \caption{
    \textbf{Native frontier geometry and coordinate projections.}
    \textbf{A}: Schematic native geometry for a fixed regularizer $f$.
    The lower curve gives the attainable-loss frontier in the
    $(I_f(S;A),L)$ plane; channels lie on or above this curve. The upper
    curve is the matched hedge-certificate frontier. The red segment marks
    the certificate gap at one operating point.
    \textbf{B}: Categorical toy problem with the KL, Pearson $\chi^2$, and
    squared Hellinger lower and certificate curves evaluated in the Shannon
    coordinate $I(S;A)$.
    \textbf{C}: The same channels evaluated in the Pearson coordinate
    $I_{\chi^2}(S;A)$. Projection changes only the horizontal coordinate.
    }
    \label{fig:frontier-geometry}
\end{figure}

\paragraph{The native rate-distortion frontier.}
We start by characterizing the Pareto frontier of attainable-losses
given an information geometry.
Fix $P(S)$, the loss $\ell(s,a)$, and the regularizer $f$. Analogous
to rate-distortion theory \citep{Shannon1959,Berger1971,ZakaiZiv1975,BenTalTeboulle1986}, we
define the smallest loss attainable with at most~$R$ units of native
information as
\begin{equation}
    \label{eq:f-rd}
    D_f(R)
    :=
    \inf_{P(A|S)}
    L,
    \qquad
    \text{s.t. } I_f(S;A)\le R.
\end{equation}
Equivalently, we can state this as an optimization of the information $R_f$ for a loss budget $D$:
\[
    R_f(D)
    :=
    \inf_{P(A|S)}
    I_f(S;A),
    \qquad
    \text{s.t. } L\le D.
\]
Every channel lies on or above this curve. Indeed, if a channel has
$I_f(S;A)\le R$, then its loss is one of the values over which the
infimum in~\eqref{eq:f-rd} is taken. Hence no decision maker evaluated
on the same task with the same regularizer can fall below $D_f(R)$ in
the $(I_f,L)$ plane.

It turns out that a bounded-rational decision maker lies on this
Pareto frontier. The reason is that the bounded-rational objective
\begin{equation}
    \label{eq:f-br-avg}
    \inf_{P(A|S)}
    \left\{
        L+\frac{1}{\beta}I_f(S;A)
    \right\}
\end{equation}
is the Lagrangian parametrization of~\eqref{eq:f-rd}
\citep{Berger1971,BenTalTeboulle1986}. If the channel selected at
$\beta$ were not on the lower frontier, another channel would have no
larger native information use and strictly smaller loss. That channel
would also have a strictly smaller value of the Lagrangian, 
contradicting optimality. Thus, for fixed $f$,
varying $\beta$ traces the lower attainable-loss frontier, up to flat
portions or nonunique optimal channels.

\paragraph{Native lower and certificate curves.}
For each $\beta>0$, let $P(A|S)$ be a channel selected by the
bounded-rational rule, and write $I_f(\beta)$, $L(\beta)$, and
$L_f^{\mathrm{adv}}(\beta)$ for the corresponding native information,
loss, and certificate value. Since the bounded-rational objective is
the Lagrangian parametrization of the native rate-distortion problem
\citep{Berger1971,BenTalTeboulle1986}, the optimizers define the
native lower curve
\begin{equation}
    \label{eq:trade-off-curve}
    \Gamma_f
    :=
    \bigl\{ (I_f(\beta),L(\beta)) : \beta>0 \bigr\}.
\end{equation}
This curve lies on the Pareto frontier described by $D_f(R)$. The same selected channels
also generate the matched certificate curve
\begin{equation}
    \label{eq:adv-curve}
    \Gamma_f^{\mathrm{adv}}
    :=
    \bigl\{ (I_f(\beta),L_f^{\mathrm{adv}}(\beta)) : \beta>0 \bigr\},
\end{equation}
where $L_f^{\mathrm{adv}}(\beta)$ is defined in
\eqref{eq:certificate}. The two curves are paired pointwise by
\[
    L_f^{\mathrm{adv}}(\beta)-L(\beta)
    =
    \frac{1}{\beta}I_f(\beta).
\]
The lower curve is the native Pareto frontier, and the upper curve is
its matched certificate frontier.

\paragraph{Comparing decision rules.}
Different regularizers define different native planes. To compare
decision rules governed by different regularizers, choose a reference
regularizer $g$ and re-evaluate the same induced channels in that
reference coordinate. Concretely, a channel selected by the
$f$-regularized rule is plotted with horizontal coordinate
\[
    I_g(S;A)
    = D_g\bigl(P(S,A)\|P(S)P(A)\bigr),
\]
using the same joint law $P(S,A)$ and the same marginal $P(A)$ induced
by that channel. Thus a point represented in the native $f$-plane as
$(I_f(S;A),L)$ is plotted in the $g$-plane as $(I_g(S;A),L)$. This
change of information coordinate keeps the loss fixed, but
can stretch, compress, bend, or cross loss curves because the
horizontal displacement is determined by the geometry of the target
$g$-plane. In it, the $g$-regularized loss curve is the  
Pareto frontier: at a fixed
loss level, it uses the smallest attainable value of $I_g(S;A)$.
Projected $f$-curves are therefore evaluated by their efficiency under
the reference information coordinate, and Pareto optimality in the
native $f$-plane need not be preserved after projection
\citep{LieseVajda2006,SasonVerdu2018}. The same construction applies
to certificates: the $f$-certificate is plotted as
$(I_g(S;A),L_f^{\mathrm{adv}})$, with the hedge still determined by
$f$. Hence the vertical certificate gap remains the $f$-native hedge in
loss units, while areas between projected curves depend on the chosen
horizontal coordinate and on the operating range. Common-plane plots
therefore compare response laws under a shared information coordinate,
while the lower frontier and certificate frontier remain native to the
regularizer that generated them.

\paragraph{KL and rate-distortion.}
For KL, \eqref{eq:f-br-avg} becomes
\begin{equation}
    \label{eq:itbr-curve}
    \int P(s)\,F_\beta[P(A|s)]\,ds
    = L + \frac{1}{\beta}\,I(S;A).
\end{equation}
Multiplying by $\beta$ yields
\[
    I(S;A) + \beta L,
\]
which is the Lagrangian formulation of Shannon's
rate-distortion problem \citep{Shannon1959,Berger1971}. At operating level $\beta$, the minimizing
channel has the Gibbs form
\begin{equation}
    \label{eq:gibbs}
    P(a|s) = \frac{1}{Z(s)}\,P(a)\,e^{-\beta \ell(s,a)},
\end{equation}
where $Z(s)$ is the normalizing constant and the marginal satisfies the
self-consistency condition~\eqref{eq:marginal-self-consistency}
\citep{Blahut1972,Arimoto1972}. Together,
\eqref{eq:marginal-self-consistency} and \eqref{eq:gibbs} are the 
Blahut--Arimoto fixed-point equations that characterize optimal channels
\citep{Blahut1972,Arimoto1972}.
As $\beta$ varies, the corresponding optimal channels trace the
native lower frontier in the KL $(I,L)$ plane. This is what
Figure~\ref{fig:frontier-geometry}C--D shows: the KL operating curve
lies on the Shannon rate-distortion frontier.

\section{Estimating the Frontier from Data}
\label{sec:estimation}

The preceding sections define the lower frontier and the certificate
frontier for a fixed response law. We now describe how these objects are
estimated from black-box behavior. The empirical goal is to recover,
along the explored operating range,
\[
    L,
    \qquad
    L_f^{\mathrm{adv}},
    \qquad
    I_f(S;A),
    \qquad
    C_s^{\mathrm{opt}}(a).
\]

\begin{figure}[!tb]
    \centering
    \includegraphics[width=\linewidth]{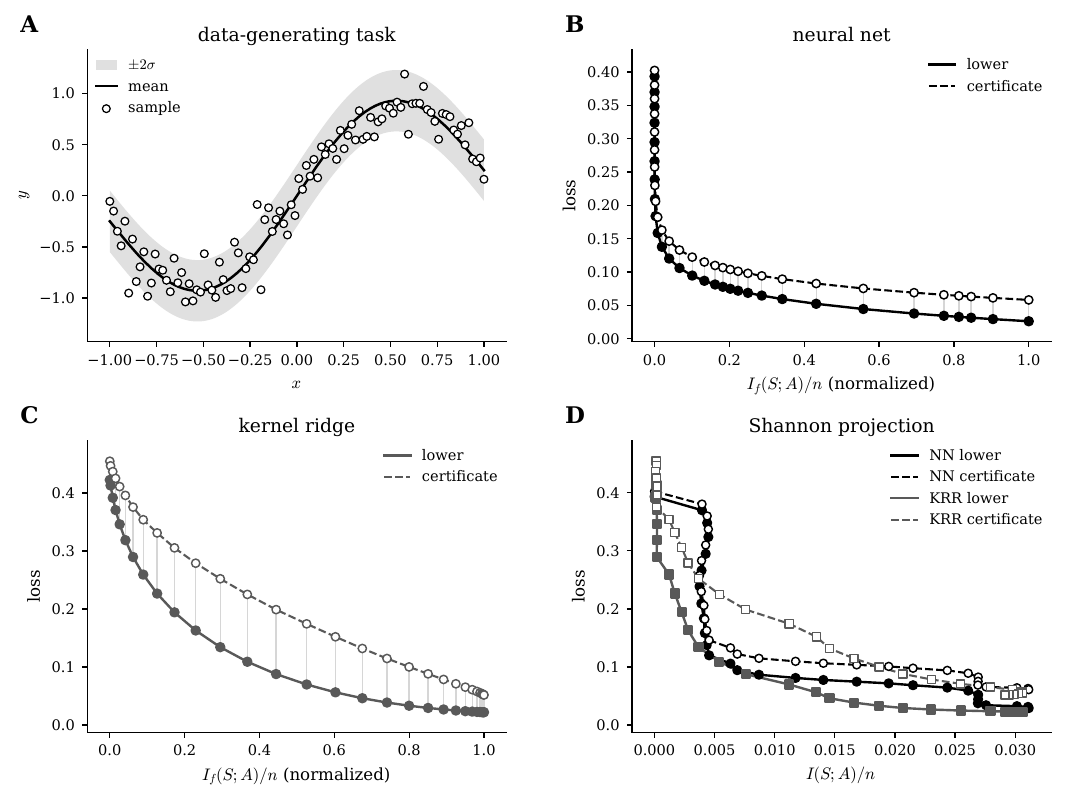}
    \caption{
    \textbf{Estimating lower and certificate curves for black-box learners.}
    \textbf{A:} Regression task, with the mean function,
    a $\pm 2\sigma$ noise band, and one sampled training set.
    \textbf{B:} Two-layer ReLU network with one hidden layer
    of 32 units, trained on squared loss with training time
    as the operating control. The solid and dashed
    curves are the estimated lower (loss) and certificate curves,
    respectively, both shown in the learner's native information
    coordinate.
    \textbf{C:} Kernel ridge regression with a Gaussian kernel, 
    with the ridge parameter as the operating control, 
    estimated in the same way.
    \textbf{D:} Projection of the same operating points into the Shannon
    information coordinate. The losses and certificates are unchanged;
    only the horizontal coordinate is replaced by $I(S;A)/n$.
    Light bands indicate bootstrap uncertainty over sampled training sets.
    See Appendix~\ref{app:experiment} for details.
    }
    \label{fig:nominal-frontier}
\end{figure}

The estimation uses three interventions. Repeated stimulus--action
observations estimate the channel and hence the operating loss. Scaling
the loss estimates the average certificate. Varying an admissible
operating control then recovers the native information coordinate along
the path. Figure~\ref{fig:nominal-frontier} illustrates this procedure
for two black-box learners on the same regression task. Local loss
perturbations, discussed below, recover the hedge itself.

\paragraph{Separability and admissible operating paths.}
A structural assumption at the heart of the free energy objective is
the separation between the
regularizer $f$ and the operating parameter $\beta$. The regularizer
$f$ determines how to measure the dependence between $S$ and $A$ and which
hedge produces the certificate. The parameter $\beta$ sets the exchange
rate between loss and native information use for that fixed response
rule. Thus, for fixed $f$, increasing $\beta$ makes native information
use less costly. A bounded-rational optimizer therefore moves
along the corresponding information--loss tradeoff curve, with
$I_f(S;A)$ non-decreasing and loss~$L$ non-increasing.

Empirically, this separation determines which observed curves can be
interpreted as operating paths. An \emph{admissible operating path} is a
sequence of observed channels produced by varying a control such as
training time, optimization budget, numerical precision, or an
annealing schedule within a fixed algorithmic family, in a regime where
the same regularizer governs the response and native information use
increases monotonically. Under these conditions the observed channels
are different operating points of one bounded-rational tradeoff curve.
Controls that alter the effective response law, such as changing the
architecture class, optimizer, augmentation scheme, noise model, or
regularization, produce observations from different response regimes.

\paragraph{Operating loss.}
At a fixed operating setting, repeated observations of $(s,a)$ estimate
the induced channel $P(A|S)$ and the marginal
$P(a)=\int P(a|s)P(s)\,ds$. The operating loss is
\begin{equation}
    \label{eq:estimate-loss}
    L
    =
    \int\!\!\int P(s,a)\ell(s,a)\,da\,ds.
\end{equation}
Varying an operating control gives a curve of observed losses. Such a
curve represents a single bounded-rational operating path only when the
control changes the strength of sample dependence while preserving the
same response law. 

\paragraph{Average certificate.}
The average certificate can be estimated without fitting an
$f$-divergence model. Fix one operating setting and replace the loss by
$t\ell(s,a)$, with~$0\le t\le 1$. For each value of~$t$, run the same
black box at the same operating setting and evaluate the induced channel
using the original loss~$\ell$. Let this value be $L(t)$, so
$L(1)=L$.

For a bounded-rational response along this scaled-loss path, the
first variation gives
$t\,\delta L+(1/\beta)\delta I_f(S;A)=0$. Integrating from $t=0$ to
$t=1$ shows that the information cost in loss units is the accumulated
loss decrease along the path. Hence the certificate value is estimated
directly by
\begin{equation}
    \label{eq:estimate-certificate}
    L_f^{\mathrm{adv}}
    = \int_0^1 L(t)\,dt.
\end{equation}
See Appendix~\ref{app:loss-scaling}.
In practice this integral is computed by numerical quadrature over the
loss-scaling experiments. Repeating the same experiment at each
operating setting gives the empirical certificate curve. The
certificate gap is then $L_f^{\mathrm{adv}}-L = \frac{1}{\beta}I_f(S;A)$.

\paragraph{Native information coordinate.}
The certificate gap gives the information cost in loss units. To recover
the native information coordinate itself, use the operating path. Along
a bounded-rational path, $dI_f(S;A)=-\beta\,dL$, while
$I_f(S;A)=\beta(L_f^{\mathrm{adv}}-L)$. Combining the two identities
yields
\begin{equation}
    \label{eq:estimate-beta-info}
    d\log\beta
    =
    -
    \frac{dL_f^{\mathrm{adv}}}{L_f^{\mathrm{adv}}-L},
    \qquad
    I_f(S;A)
    =
    \beta\bigl(L_f^{\mathrm{adv}}-L\bigr).
\end{equation}
This is derived in Appendix~\ref{app:path-identity}.
Thus the observed loss curve and certificate curve determine $\beta$
along the operating path up to one multiplicative constant. This
constant fixes the unit of native information. Indeed, replacing $f$ by
$cf$ and $\beta$ by $c\beta$ leaves the response and the certificate gap
unchanged, while rescaling $I_f(S;A)$ by $c$. Once this scale is fixed,
\eqref{eq:estimate-beta-info} gives the native information coordinate
along the observed path.

\paragraph{Local hedge.}
Loss scaling identifies the average certificate. To identify the local
hedge, perturb the loss at a fixed operating setting from $\ell(s,a)$ to
$\ell(s,a)+\epsilon C(s,a)$ and observe the first-order change in the
channel and marginal. The observable object is the cost correction that
makes the actions used by the observed channel locally indifferent.
Thus, for actions in the support of $P(a|s)$,
\begin{equation}
    \label{eq:estimate-local-hedge}
    C_s^{\mathrm{opt}}(a)-C_s^{\mathrm{opt}}(a')
    =
    \ell(s,a')-\ell(s,a).
\end{equation}
Off the support, the corresponding inequality is the one in
\eqref{eq:indifference}. The hedge is determined by
\eqref{eq:estimate-local-hedge} up to stimulus-wise constants, since
adding the same value to all actions at fixed $s$ does not change the
response. The aggregate constant is fixed by the certificate gap:
\[
    \int\!\!\int P(s,a)C_s^{\mathrm{opt}}(a)\,da\,ds
    =
    L_f^{\mathrm{adv}}-L.
\]
Small perturbations of the loss then reveal how this recovered
correction changes as the channel moves. An $f$-divergence model is
compatible on the explored range when one regularizer explains these
local corrections and their perturbation responses across operating
points.

\section{Generalization as Hedging Against Sample Distortion}
\label{sec:generalization}

Learning is bounded-rational acting with samples as stimuli and
predictors as actions. Generalization enters this picture through
sample distortion: the learner chooses a predictor guided by empirical
loss, but performance is evaluated by population loss. The discrepancy
between these two losses acts as an adversarial perturbation.

Throughout this section, we work in the standard sample-to-population
setting: the training observations are drawn i.i.d.\ from the same
population distribution that defines population loss. As shown below,
this assumption makes the sample distortion centered under the product
law. The general hedging framework also applies to distribution shift,
deployment mismatch, and model misspecification, where the evaluation
perturbation need not be centered.

\paragraph{Learning as bounded-rational acting.}
Let $S=(Z_1,\dots,Z_n)$ be the training sample, where the observations
$Z_i$ are drawn i.i.d.\ from $P(Z)$, and let $A$ be a random learned
predictor. Since $n$ sets the natural scale of the problem, we use
per-observation losses and native information rates throughout.
For realizations $S=s$ and $A=a$, write
\begin{equation}
    \label{eq:train-pop-loss}
    \ell_n(s,a)
    :=
    \frac{1}{n}\sum_{i=1}^n \ell(z_i,a)
    \qquad\text{and}\qquad
    \bar L(a)
    :=
    \int \ell(z,a)\,P(z)\,dz
\end{equation}
for the empirical and population losses. The channel loss is
\[
    L_n
    :=
    \int\!\!\int P(s,a)\,\ell_n(s,a)\,da\,ds.
\]
For a chosen regularizer $f$, the bounded-rational learning objective is
\begin{equation}
    \label{eq:learning-free-energy}
    F_{\beta,n}[P(A|S)]
    =
    L_n
    +
    \frac{1}{\beta n}\,
    I_f(S;A).
\end{equation}
Thus, at fixed $n$, the learner operates in the native plane
$(\tfrac{1}{n}I_f(S;A),L_n)$. Bounded-rational optimization
selects supporting points of the lower frontier in this plane: if
another channel had no larger native information rate and strictly
smaller empirical loss, it would strictly reduce
\eqref{eq:learning-free-energy}. Varying $\beta$ at fixed $n$ changes
how strongly the sample controls the output.

\paragraph{Sample distortion.}
The learning-specific perturbation is the discrepancy between
population loss and empirical loss. For a sample $s$, define the sample
distortion by
\begin{equation}
    \label{eq:sample-distortion}
    C_{s,n}(a)
    :=
    \bar L(a)-\ell_n(s,a).
\end{equation}
Since the observations in $S$ are i.i.d., the expected empirical loss
equals the population loss:
\[
    \int P(s)\ell_n(s,a)\,ds
    =
    \frac{1}{n}\sum_{i=1}^n
    \int \ell(z_i,a)\,P(z_i)\,dz_i
    =
    \bar L(a).
\]
Therefore the expected distortion is zero under the product law:
\begin{equation}
    \label{eq:distortion-product-zero}
    \int\!\!\int P(s)P(a)C_{s,n}(a)\,da\,ds
    = \int P(a)
    \left[\bar L(a)-\int P(s)\ell_n(s,a)\,ds\right]\,da
    = 0.
\end{equation}
This centering property distinguishes sample distortion from an ordinary
decision-making perturbation: before the learner conditions its output on
the sample, the perturbation has no average effect.

After conditioning, the same perturbation is evaluated under the
learner's joint law. Since $\bar L(a)=\ell_n(s,a)+C_{s,n}(a)$,
population loss decomposes as
\begin{equation}
    \label{eq:population-as-distorted-training}
    \int\!\!\int P(s,a)\,\bar L(a)\,da\,ds
    = L_n + \int\!\!\int P(s,a)\,C_{s,n}(a)\,da\,ds,
\end{equation}
where the second term is the channel-weighted sample distortion.
Unlike \eqref{eq:distortion-product-zero}, this term is evaluated under
the joint law induced by the learner. Using
$P(s,a)=P(s)P(a|s)$ and \eqref{eq:distortion-product-zero}, it can be
written as
\begin{equation}
    \label{eq:gap-as-reweighting}
    \int\!\!\int P(s,a)C_{s,n}(a)\,da\,ds
    = \int\!\!\int P(s)P(a)
    \left[
        \frac{P(a|s)}{P(a)}-1
    \right]
    C_{s,n}(a)\,da\,ds.
\end{equation}
This identity isolates the role of the channel. The sample distortion
has zero average under the product law $P(S)P(A)$, and it becomes
population cost only through the likelihood-ratio displacement
$P(a|s)/P(a)-1$ induced by the learner.

Overfitting is the amplification of sample distortion by sample
dependence. Moving down the lower frontier reduces empirical loss by
allowing the sample to redirect the output more strongly. Equation
\eqref{eq:gap-as-reweighting} gives the population consequence of that
movement. If the likelihood-ratio displacement assigns extra mass to
actions with positive $C_{s,n}(a)$, then the channel increases
population loss relative to empirical loss. If it assigns extra mass to
actions with negative $C_{s,n}(a)$, then the same dependence lowers
population loss relative to empirical loss.

\paragraph{The endogenous hedge.}
The bounded-rational rule supplies a matched hedge in the same loss
coordinate. At fixed $n$, the hedge induced by the regularizer is
\begin{equation}
    \label{eq:learning-opt-perturbation}
    C_{s,n}^{\mathrm{opt}}(a)
    =
    \frac{1}{\beta n}
    \left[
        f'\!\left(\frac{P(a|s)}{P(a)}\right)
        +
        G(a)
    \right],
\end{equation}
and the corresponding certificate value is
\begin{equation}
    \label{eq:learning-certificate}
    \begin{aligned}
    L_f^{\mathrm{adv}}(\beta)
    &:=
    \int\!\!\int P(s,a)
    \left[
        \ell_n(s,a)+C_{s,n}^{\mathrm{opt}}(a)
    \right]da\,ds  \\
    &=
    L_n+\frac{1}{\beta n}I_f(S;A).
    \end{aligned}
\end{equation}
The equality follows from the certificate identity in learning units:
the average hedge equals the native information cost converted into
loss units.

\paragraph{Certificate condition.}
The certificate compares the native hedge with the actual sample
distortion under the learner's own channel. Combining
\eqref{eq:population-as-distorted-training} and
\eqref{eq:learning-certificate} gives the exact margin
\begin{equation}
    \label{eq:learning-certificate-margin}
    L_f^{\mathrm{adv}}(\beta)
    -
    \int\!\!\int P(s,a)\bar L(a)\,da\,ds
    =
    \int\!\!\int P(s,a)
    \left[
        C_{s,n}^{\mathrm{opt}}(a)-C_{s,n}(a)
    \right]da\,ds.
\end{equation}
Therefore population loss is below the certificate,
\begin{equation}
    \label{eq:population-below-learning-certificate}
    \int\!\!\int P(s,a)\,\bar L(a)\,da\,ds
    \le L_f^{\mathrm{adv}}(\beta),
\end{equation}
if and only if the right hand side of
\eqref{eq:learning-certificate-margin} is nonnegative. The regularizer
determines both quantities in this comparison: it determines the channel
that weights the sample distortion, and it determines the hedge whose
surplus is evaluated.

The margin gives the generalization certificate empirical content. The
operating loss and certificate curve are recovered from loss-scaling
experiments, while local perturbations identify
$C_{s,n}^{\mathrm{opt}}(a)$ up to stimulus-wise constants fixed by the
certificate gap. Evaluation against the data-generating distribution, or
against held-out samples from it, estimates $\bar L(a)$ and hence
$C_{s,n}(a)$. The recovered hedge certifies an operating point exactly
when the channel-weighted surplus in
\eqref{eq:learning-certificate-margin} is nonnegative.

\paragraph{The role of sample size.}
The sample size fixes the per-observation scale of the geometry.
Information use is measured by $\tfrac{1}{n}I_f(S;A)$, and the hedge is
scaled by $\tfrac{1}{\beta n}$. Changing $\beta$ moves the learner along
the fixed-$n$ native frontier. Changing $n$ changes the empirical loss,
the sample distortion, and the corresponding family of native
frontiers.

\section{Three Consequences}
\label{sec:consequences}

The following consequences are three readings of the certificate
identity \eqref{eq:learning-certificate-margin}. They expose a blind
spot in scalar diagnostics of generalization: empirical loss, information
use, and perturbation size do not determine whether the learner's
response law covers the distortion it faces, because the hedge depends on the
full channel itself.

\paragraph{Dangerous samples are hedge-deficit samples.}

Dangerous samples are often identified by hard-example criteria or by
influence diagnostics, such as high loss, difficult margins, or large
effect on predictions \citep{ShrivastavaGuptaGirshick2016,KohLiang2017}.
The certificate gives a different diagnostic. A sample is dangerous
when its distortion exceeds the learner's hedge on the actions the
learner actually assigns probability to:
\[
    \int P(a|s)
    \bigl[
        C_{s,n}(a)-C_{s,n}^{\mathrm{opt}}(a)
    \bigr]\,da .
\]
Averaged over samples, this is the samplewise contribution to
certificate failure. Thus empirical difficulty and sample danger need
not agree. A low-loss sample can be dangerous when it redirects the
learner toward actions whose population loss is worse than the hedge
covers. A high-loss sample can be harmless when its distortion is
covered by the hedge.

The mechanism follows from \eqref{eq:population-as-distorted-training}.
The sample affects population loss through the channel-weighted
distortion term, not through empirical loss alone. Dataset auditing,
poisoning detection, reweighting, and active data collection should
therefore rank samples by hedge deficit when the goal is to identify
certificate failure.

\paragraph{Equal-sized perturbations can have opposite effects.}

Distributional robustness and domain adaptation often measure shift by
a distance from the training distribution, such as KL, Wasserstein
distance, or a norm \citep{EsfahaniKuhn2018,DuchiNamkoong2021}.
The certificate shows that perturbation size alone is insufficient.
Equal-sized centered perturbations can have opposite effects when they
move the loss in different directions relative to the learner's channel.

If the evaluation perturbation changes by a centered amount
$\delta C_s(a)$, with $\int P(s)\delta C_s(a)\,ds=0$ for every action,
then \eqref{eq:learning-certificate-margin} changes by
$-\int\!\!\int P(s,a)\delta C_s(a)\,da\,ds$. To see the effect, consider
a two-stimulus, two-action task with $P(s_1)=P(s_2)=1/2$. Let the learner
put probability $q>1/2$ on $a_1$ at $s_1$ and on $a_2$ at $s_2$.
Define a centered perturbation by assigning values $c,-c$ to
$(s_1,s_2)$ for action $a_1$, and values $-c,c$ for action $a_2$.
Its negative has the same size under any symmetric norm. Yet the
channel-weighted effects are opposite:
\[
    \int\!\!\int P(s,a)\delta C_s(a)\,da\,ds
    =
    c(2q-1),
    \qquad
    \int\!\!\int P(s,a)(-\delta C_s(a))\,da\,ds
    =
    -c(2q-1).
\]
Thus two equal-sized centered perturbations can have opposite effects on
the certificate margin. The sign is determined by alignment with the
learner's channel.

This matters in deployment because target labels are often unavailable
or delayed. A scalar shift distance ranks environments by magnitude.
The hedge ranks them by how their induced loss perturbations interact
with the response law. The relevant quantity is therefore the
channel-weighted perturbation, not the perturbation size alone.

\paragraph{Less sample dependence need not mean better generalization.}

Machine learning theory often treats reduced sample dependence as
favorable: smaller information use, stronger compression, heavier
regularization, or a simpler effective model are expected to improve
generalization \citep{McAllester1999,XuRaginsky2017}. The
bounded-rational view gives a more precise criterion. Sample dependence
is harmful when it amplifies positive sample distortion, and useful
when it moves probability mass toward actions with negative or covered
distortion.

The mechanism is the reweighting identity
\eqref{eq:gap-as-reweighting}. The native information $I_f(S;A)$
measures the size of the channel displacement in the regularizer's
geometry. It does not determine the sign of the distortion term. Thus
reducing $I_f(S;A)$ can remove useful dependence, while increasing it
can improve population loss when the additional dependence aligns
favorably with $C_{s,n}(a)$.

This consequence subsumes several common interventions. Pruning can
preserve empirical loss while removing actions that carry hedging
value. Aggressive regularization can reduce sample dependence while
weakening protection against the relevant distortion. Label smoothing
can improve calibration or the smoothed objective while moving
probability mass into directions not covered by the original-task
hedge. In each case the decisive quantity is the effect of the
intervention on the channel-weighted distortion and on the recovered
hedge, not the reduction of sample dependence alone.

\section{Related work}
\label{sec:related}

\paragraph{Bounded-rational response laws.}
The starting point of this paper is information-theoretic bounded
rationality, where a decision maker trades expected loss against the
cost of changing an action distribution after observing a stimulus
\citep{OrtegaBraun2011,OrtegaBraun2013,BraunOrtega2014,OrtegaBraunDyerKimTishby2015}.
This framework gives the free-energy decision rule used in
Section~\ref{sec:bounded-rationality}. Empirical work on human choice
under time pressure supports the interpretation of the operating
parameter as a resource variable controlling how strongly observations
redirect actions \citep{OrtegaStocker2016}. Rational-inattention models
give a closely related revealed-preference view of information-constrained
choice \citep{Sims2003,MatejkaMcKay2015,CaplinDean2015}. This paper
takes the response law as the primitive object and uses black-box
behavior to recover the regularizer-native hedge, the lower
attainable-loss frontier, and the certificate frontier.

\paragraph{Generalized information coordinates.}
The use of $f$-divergences follows the classical theory of generalized
statistical divergence
\citep{Morimoto1963,AliSilvey1966,Csiszar1967,Vajda1968}. Csiszár's
informativity functionals and related treatments show that channels can
be evaluated by dependence measures other than Shannon mutual information
\citep{Csiszar1972,ZivZakai1973,Csiszar1995,LieseVajda2006,SasonVerdu2018}.
These works provide the provenance for the native coordinate
$I_f(S;A)$. We use $f$-divergences as a flexible and mathematically
simple extension of KL regularization to illustrate the concept of
innate regularization.
The selected $f$ determines the dependence
coordinate in which the learner's response is evaluated. It therefore
also determines the native information--loss geometry. Cross-regularizer
comparison in a Shannon plane is a projection of the induced channels
into a common coordinate system.

\paragraph{Rate-distortion and lower frontiers.}
The lower frontier is inherited from rate-distortion theory. Shannon's
formulation and Berger's development identify the tradeoff between
distortion and mutual information in the KL case
\citep{Shannon1959,Berger1971}, with the Blahut--Arimoto equations
giving the corresponding fixed-point characterization
\citep{Blahut1972,Arimoto1972}. Generalized rate-distortion theory
extends this frontier construction to generalized information measures
\citep{ZakaiZiv1975,BenTalTeboulle1986}, and recent work studies
$f$-information in lossy compression and generalization
\citep{MasihaGohariYassaee2023}. This paper uses the lower-frontier
geometry as one side of a paired construction. The same regularizer also
generates an upper certificate frontier, so attainable loss and hedged
loss are described in one native geometry.

\paragraph{Duality, hedging, and robust control.}
The hedging interpretation comes from convex duality
\citep{Fenchel1949,Rockafellar1970,BenTalTeboulle1987}. Information-theoretic
bounded rationality admits an adversarial interpretation:
the information cost is the price of protection against loss
perturbations \citep{OrtegaLee2014}. Related dualities appear in
risk-sensitive and robust control, where exponential criteria and
worst-case perturbations are linked
\citep{HowardMatheson1972,Jacobson1973,Whittle1990,HansenSargent2001,HansenSargent2008}.
Regularized Markov decision processes and reinforcement-learning
formulations develop analogous relations between policy regularization
and robustness
\citep{GeistScherrerPietquin2019,HusainCiosekTomioka2021,DermanGeistMannor2021,EysenbachLevine2022}.
The closest reinforcement-learning analogue shows that a policy
regularizer implicitly defines an adversarial reward 
perturbation~\citep{BrekelmansGeneweinGrauMoyaDeletangKuneschLeggOrtega2022}. 
This paper develops the corresponding channel-level hedge for
bounded-rational acting with $f$-native regularizers and uses it to
construct an empirically recoverable certificate curve.

\paragraph{Distributionally robust optimization (DRO).}
Distributionally robust optimization studies decisions evaluated under
sets of plausible probability laws. These sets may be defined by
moments, divergence balls, Wasserstein neighborhoods, or other
analyst-specified uncertainty models
\citep{DelageYe2010,GohSim2010,BenTalDenHertogDeWaegenaereMelenbergRennen2013,HuHong2013,WiesemannKuhnSim2014,EsfahaniKuhn2018,RahimianMehrotra2022}.
Divergence-based DRO is especially close in form because it uses
$\phi$- or $f$-divergence penalties to define robust neighborhoods
\citep{BayraksanLove2015,LoveBayraksan2016,Lam2016,DuchiNamkoong2021}.
DRO begins with an externally specified ambiguity set. This paper begins
with the observed response law. The perturbation geometry is induced by
the bounded-rational rule itself and is recovered from the learner's
behavior. The resulting certificate is therefore native to the observed
decision maker.

\paragraph{Learning and generalization.}
PAC-Bayesian theory and information-theoretic generalization also study
the dependence between the training sample and the learned output.
PAC-Bayes controls population performance through a comparison between a
posterior and a reference distribution
\citep{McAllester1999,Seeger2002,Catoni2007}. Empirical PAC-Bayes work
by Dziugaite and Roy shows that such bounds can be made nonvacuous for
modern stochastic neural networks and that data-dependent priors can be
handled under additional validity conditions
\citep{DziugaiteRoy2017,DziugaiteRoy2018}. Information-theoretic analyses
bound generalization through mutual information, individual information,
or conditional mutual information between sample and output
\citep{RaginskyRakhlinTsaoWuXu2016,XuRaginsky2017,BuZouVeeravalli2020,SteinkeZakynthinou2020,HaghifamNegreaKhistiRoyDziugaite2020};
a synthesis is given by
\citet{HellstromDurisiGuedjRaginsky2024}. These works derive upper
bounds on the generalization gap under analyst-specified assumptions.
This paper derives its learning statement from the bounded-rational
hedge. Population loss is empirical loss plus sample distortion, and the
certificate applies exactly when the recovered native hedge covers that
distortion, as expressed in
\eqref{eq:learning-certificate-margin}. The learning contribution is
therefore a practical hedging test for the learner's own response law.

\section{Discussion}
\label{sec:discussion}

This paper has treated generalization as a property of the learner's
own response law. The learner's dependence on the sample induces both
exposure to sample distortion and a native hedge against that exposure.
Classical analyses typically ask how large the generalization gap can be
for a given sample-to-output channel under an analyst-chosen control,
such as a prior, a stability condition, a complexity measure, or a tail
assumption \citep{HellstromDurisiGuedjRaginsky2024}. In contrast, the present
construction asks whether the learner's recovered native hedge \emph{covers}
the sample distortion encountered at the operating point being tested.
The precise statement is \eqref{eq:learning-certificate-margin}:
population loss is below the certificate exactly when the
channel-weighted surplus of the native hedge over sample distortion is
nonnegative. Thus the relevant object is the learner's own hedge against
the distortion selected by its own dependence on the sample.

The native hedge is found through the stable response behavior across
interventions. Loss scalings, local perturbations, and operating changes
constrain which regularizer can govern the learner's response over an
operating range. A single channel may admit many regularized
representations, but a stable response geometry must also satisfy the
indifference relations in \eqref{eq:indifference}, the certificate
identity in \eqref{eq:learning-certificate}, and the path relation in
\eqref{eq:estimate-beta-info}. When one regularizer explains these
restrictions over the explored range, the resulting hedge is tied to
what the learner actually does rather than to an externally supplied
prior or complexity measure.

Thus, the comparison with classical generalization analyses basically
reduces to a comparison of distortion control mechanisms. 
Classical bounds control the sample-to-output
channel through assumptions chosen by the analyst. The costruction here
recovers the control from the learner's behavior and evaluates the
sample distortion against that recovered control. The two approaches
answer complementary questions: external assumptions give guarantees
that hold uniformly under the analyst's model, while the native hedge
gives an operating-point certificate for the learner's observed response
law. They can therefore be combined by using external assumptions to
control tails while preserving the regularizer-native hedge as the
source of the certificate.

The certificate applies under testable conditions. First, it is an
expected-value certificate for the channel-weighted distortion. A
high-probability statement requires a separate tail step, such as the
binary $f$-divergence transfer in \eqref{eq:bindiv-mi}, together with a
product-law tail estimate for the perturbation. Second, recovery is
valid on ranges where the response geometry is stable. Residuals in the
indifference relations, the certificate identity, and the path relation
provide empirical evidence for or against this stability over the
explored range. Third, an operating path must remain within one
regularizer regime. Controls that alter the effective response law
produce observations from a different regime and should be analyzed as a
different native geometry.

The next step is to combine the expected certificate with native tail
transfer while preserving the same regularizer-native geometry. This
would turn the recovered hedge into a high-probability statement without
replacing it by an external information coordinate. The most immediate
empirical problem is to identify operating controls that preserve
response geometry across scale. Once such controls are available, the
learner's sample dependence can be evaluated in its own native geometry,
making generalization an empirically testable hedging property of the
observed response law.

\section{Conclusions}
\label{sec:conclusions}

Thinking amplifies distortions. Conditioning on $S$ moves $A$ toward
lower loss. It also carries errors in the loss into the joint law
$P(S,A)$. The channel that improves fit also creates exposure. Bounded
rationality supplies the hedge: the regularizer determines which
perturbations are covered, and $\beta$ sets how far the situation may
redirect the response \citep{OrtegaBraun2013,
OrtegaBraunDyerKimTishby2015,OrtegaLee2014}.

Robustness is encoded in response. The primitive object is the response
law $P(A|S)$. Its behavior under loss scaling, perturbations, and
changes in operating level reveals the regularizer $f$. Once recovered,
$f$ determines the native information coordinate $I_f(S;A)$, the lower
frontier, and the certificate frontier. The certificate identity shows
that information cost and hedging premium are the same quantity in loss
units. Robustness is therefore revealed by behavior.

Generalization is internal hedging. The response law determines both exposure
to distortion and the hedge against it. This is where acting and
learning meet: in both cases, the response is formed under one loss and
evaluated under another. In the classical sample-to-population case, this
perturbation is centered under the product law, as in
\eqref{eq:distortion-product-zero}, so the generalization gap is the
channel-reweighting effect in \eqref{eq:gap-as-reweighting}. The
hedging logic goes beyond this case. Under distribution shift, 
deployment mismatch, or an imprecise model of the situation,
the certificate condition remains the same: the recovered
hedge must cover the evaluation perturbation under the joint law, as
expressed in \eqref{eq:learning-certificate-margin}.

Bounded rationality becomes a theory of which distortions a decision
maker can absorb. In machine learning, grokking
\citep{PowerBurdaEdwardsBabuschkinMisra2022,LiuMichaudTegmark2023},
double descent
\citep{BelkinHsuMaMandal2019,NakkiranKaplunBansalYangBarakSutskever2021},
and learning-rate schedules
\citep{LoshchilovHutter2017,CohenKaurLiKolterTalwalkar2021} could become
problems of how response geometry changes and which distortions the
resulting hedge can absorb. In reinforcement learning, reward
misspecification, simulator bias, and preference-model error can be
treated as tests of whether the policy's hedge covers the relevant loss
perturbation. In economics, limited attention and bounded rationality
reveal both an information cost and a robustness class
\citep{Sims2003,MatejkaMcKay2015,CaplinDean2015}. The common empirical
task is to recover the response geometry of real systems and identify
the distortions for which their certificates apply.
The central empirical question is no longer how much a system thinks,
but which distortions its thinking can afford.

\bibliographystyle{unsrtnat}
\bibliography{main}

\appendix

\section{Mathematical Derivations}

\subsection{Derivation of the Optimal Adversarial Perturbation}
\label{app:optimal-perturbation}

We seek the perturbation $C_s(a)$ that the adversary applies optimally.
The central claim is that this perturbation is uniquely determined by
the channel $P(A|S)$ the decision maker already operates: it equals the
functional gradient of the native information penalty $I_f(S;A)$ with
respect to the channel, converted into loss units by $1/\beta$.

The native $f$-mutual information is
\[
    I_f(S;A)
    =
    \int P(s) \int P(a)\,
    f\!\left(\frac{P(a|s)}{P(a)}\right) da\, ds,
\]
a functional of the channel $P(A|S)$. Our goal is to compute its
functional derivative $\delta I_f(S;A)/\delta P(a|s)$.
However, there is a coupling constraint.
The marginal satisfies $P(a) = \int P(s)\, P(a|s)\, ds$, so any
admissible perturbation $\delta P(a|s)$ which satisfies
$\int \delta P(a|s)\, da = 0$ for every $s$ induces
\[
    \delta P(a) = \int P(s)\, \delta P(a|s)\, ds.
\]
This coupling term must be tracked throughout.

We will now vary the integrand. Fix $(s,a)$. 
The perturbation enters $P(a)\,f(P(a|s)/P(a))$ through
$P(a|s)$ in the numerator and through $P(a)$ in the denominator and
prefactor. By the chain rule and product rule respectively, the two
contributions are
\[
    \begin{aligned}
    \text{(numerator)}&\qquad
    f'\!\left(\frac{P(a|s)}{P(a)}\right)\delta P(a|s),
    \\
    \text{(denominator)}&\qquad
    \left[
        f\!\left(\frac{P(a|s)}{P(a)}\right)
        -
        \frac{P(a|s)}{P(a)}
        f'\!\left(\frac{P(a|s)}{P(a)}\right)
    \right]\delta P(a).
    \end{aligned}
\]
The minus sign in the denominator contribution reflects that $P(a)$
appears in the denominator of the argument of $f$.
Multiplying by $P(s)$, integrating over $s$ and $a$, and substituting
$\delta P(a) = \int P(s')\,\delta P(a|s')\,ds'$ into the denominator
contribution consolidates the variation into a single integral against
$\delta P(a|s)$:
\begin{equation}
    \label{eq:info-variation}
    \delta I_f(S;A)
    =
    \int P(s) \int
    \left[
        f'\!\left(\frac{P(a|s)}{P(a)}\right) + G(a)
    \right]
    \delta P(a|s)\, da\, ds,
\end{equation}
where
\[
    G(a)
    :=
    \int P(s)
    \left[
        f\!\left(\frac{P(a|s)}{P(a)}\right)
        -
        \frac{P(a|s)}{P(a)}
        f'\!\left(\frac{P(a|s)}{P(a)}\right)
    \right] ds.
\]
The term $G(a)$ aggregates the denominator effect across all stimuli,
coupling them through the shared marginal $P(A)$. It vanishes for KL
with the canonical representative $f(x)=x\log x - x + 1$ because the
two terms inside the bracket cancel identically.
Since the variation in \eqref{eq:info-variation} is linear
in $\delta P(a|s)$, the kernel of the integral is the functional derivative:
\begin{equation}
    \label{eq:info-kernel}
    \frac{1}{P(s)}
    \frac{\delta I_f(S;A)}{\delta P(a|s)}
    =
    f'\!\left(\frac{P(a|s)}{P(a)}\right) + G(a).
\end{equation}

Now we equate two equivalent representations of the same first-order change
in the objective. On one hand, if the loss is perturbed by $C_s(a)$,
the first-order change in the objective is
\[
    \int P(s) \int \delta P(a|s)\, C_s(a)\, da\, ds.
\]
On the other hand, the same change expressed through the information
penalty is
\[
    \frac{1}{\beta}\,\delta I_f(S;A)
    = \int P(s) \int
    \delta P(a|s) \cdot
    \frac{1}{\beta P(s)}
    \frac{\delta I_f(S;A)}{\delta P(a|s)}
    \, da\, ds.
\]
Since these two expressions must agree for every admissible perturbation
$\delta P(a|s)$, their integrands must coincide pointwise, giving
\[
    C_s(a)
    = \frac{1}{\beta P(s)}
    \frac{\delta I_f(S;A)}{\delta P(a|s)}.
\]
Substituting the functional derivative \eqref{eq:info-kernel} yields
\begin{equation}
    \label{eq:opt-perturbation-derived}
    C_s^{\mathrm{opt}}(a)
    =
    \frac{1}{\beta}
    \left[
        f'\!\left(\frac{P(a|s)}{P(a)}\right)
        +
        G(a)
    \right].
\end{equation}
The first term is the direct likelihood-ratio term, scaling with the
departure of the posterior from the marginal at action $a$ under
stimulus $s$. The second term, $G(a)$, enforces consistency across
stimuli through $P(A)$. Together, the optimal perturbation is the
functional gradient of the information cost the decision maker is
already incurring, expressed in loss units via $1/\beta$. The adversary
introduces no exogenous structure; it reflects the decision maker's own
dependence on the stimulus back as an effective loss increment.

\subsection{Certificate Identity}
\label{app:certificate-identity}

By definition, the certificate gap is the channel-average value of the
optimal perturbation:
\[
    L_f^{\mathrm{adv}}-L
    =
    \int\!\!\int P(s,a)C_s^{\mathrm{opt}}(a)\,da\,ds .
\]
Using \eqref{eq:opt-perturbation}, this becomes
\[
    L_f^{\mathrm{adv}}-L
    =
    \frac{1}{\beta}
    \int\!\!\int P(s,a)
    \left[
        f'\!\left(\frac{P(a|s)}{P(a)}\right)+G(a)
    \right]da\,ds .
\]
The first contribution is
\[
    \int\!\!\int P(s)P(a|s)
    f'\!\left(\frac{P(a|s)}{P(a)}\right)da\,ds .
\]
The second contribution is $\int P(a)G(a)\,da$, which equals
\[
    \int\!\!\int P(s)P(a)
    \left[
        f\!\left(\frac{P(a|s)}{P(a)}\right)
        -
        \frac{P(a|s)}{P(a)}
        f'\!\left(\frac{P(a|s)}{P(a)}\right)
    \right]da\,ds .
\]
Adding the two contributions cancels the terms involving
$f'$, leaving
\[
    \int\!\!\int P(s)P(a)
    f\!\left(\frac{P(a|s)}{P(a)}\right)da\,ds
    =
    I_f(S;A).
\]
Hence
\[
    L_f^{\mathrm{adv}}-L
    =
    \frac{1}{\beta}I_f(S;A).
\]
For learning, the same calculation applies with empirical loss
$\ell_n(s,a)$ and exchange rate $1/(\beta n)$. Therefore
\[
    L_f^{\mathrm{adv}}(\beta)
    =
    L_n+\frac{1}{\beta n}I_f(S;A),
\]
which is \eqref{eq:learning-certificate}.

\subsection{Loss Scaling and the Average Certificate}
\label{app:loss-scaling}

Fix an operating setting and replace the loss by $t\ell(s,a)$, with
$0\le t\le 1$. Let $L(t)$ and $I_f(t)$ denote the original loss and
native information of the channel selected at scale $t$. The scaled
bounded-rational value is
\[
    V(t)
    =
    tL(t)+\frac{1}{\beta}I_f(t).
\]
At $t=0$, the loss term is absent, and the minimum information cost is
zero, attained by a channel independent of $S$. Thus $V(0)=0$.

When $t$ changes by $dt$, the direct change in the value is $L(t)dt$.
The first-order change due to the movement of the optimizing channel is
zero, because the channel at scale $t$ is stationary for the scaled
objective. Hence $dV(t)=L(t)dt$, and integrating from $0$ to $1$ gives
\[
    V(1)
    =
    \int_0^1 L(t)\,dt .
\]
Since $V(1)=L(1)+\frac{1}{\beta}I_f(1)$, the right hand side is exactly
the certificate value at the original loss scale. Therefore
\[
    L_f^{\mathrm{adv}}
    =
    \int_0^1 L(t)\,dt .
\]
This is \eqref{eq:estimate-certificate}. The same derivation explains
why $L(t)$ is evaluated with the original loss: it is the derivative of
the scaled value with respect to the loss scale.

\subsection{Native Coordinate along an Operating Path}
\label{app:path-identity}

Along an admissible operating path, the regularizer is fixed while
$\beta$ changes. On differentiable portions of the path, stationarity of
$L+\frac{1}{\beta}I_f(S;A)$ gives $dI_f(S;A)=-\beta\,dL$. Thus a
first-order loss decrease is paid for by the corresponding first-order
increase in native information.

The certificate identity gives
$L_f^{\mathrm{adv}}=L+I_f(S;A)/\beta$. Differentiating yields
\[
    dL_f^{\mathrm{adv}}
    =
    dL+\frac{1}{\beta}dI_f(S;A)
    -
    \frac{I_f(S;A)}{\beta}d\log\beta .
\]
Substituting $dI_f(S;A)=-\beta\,dL$ cancels the first two terms, so
\[
    dL_f^{\mathrm{adv}}
    =
    -
    \bigl(L_f^{\mathrm{adv}}-L\bigr)d\log\beta .
\]
Therefore
\[
    d\log\beta
    =
    -
    \frac{dL_f^{\mathrm{adv}}}{L_f^{\mathrm{adv}}-L},
    \qquad
    I_f(S;A)
    =
    \beta\bigl(L_f^{\mathrm{adv}}-L\bigr).
\]
This recovers $\beta$ along the operating path up to one multiplicative
constant, and then recovers the native information coordinate on the
same scale. The remaining scale ambiguity is intrinsic: replacing
$f$ by $cf$ and $\beta$ by $c\beta$ leaves
$L+\frac{1}{\beta}I_f(S;A)$ unchanged, while replacing $I_f(S;A)$ by
$cI_f(S;A)$.

\subsection{Binary Tail Transfer}
\label{app:tail-transfer}

Let $E=\{C_S(A)>u\}$, and let $p$ and $q$ be its probabilities
under $P(S,A)$ and $P(S)P(A)$, respectively:
\[
    p=\int_E P(s,a)\,da\,ds,
    \qquad
    q=\int_E P(s)P(a)\,da\,ds .
\]
Applying the map $(s,a)\mapsto \mathbf 1_E(s,a)$ cannot increase an
$f$-divergence, so
\[
    I_f(S;A)
    = D_f(P(S,A)\|P(S)P(A))
    \ge D_f(\mathrm{Bern}(p)\|\mathrm{Bern}(q)).
\]
If a product-law argument gives $q\le \bar q$, then either
$p\le \bar q$, or, when $p>\bar q$, monotonicity of the binary
divergence in its second argument gives
\[
    \bar q f\!\left(\frac{p}{\bar q}\right)
    + (1-\bar q)f\!\left(\frac{1-p}{1-\bar q}\right)
    \le I_f(S;A).
\]
Consequently,
\[
    P(S,A)\{C_S(A)>u\}
    \le \sup\biggl\{
        p\in[\bar q,1] \biggm|
        \bar q f\!\left(\frac{p}{\bar q}\right)
        + (1-\bar q)f\!\left(\frac{1-p}{1-\bar q}\right)
        \le I_f(S;A)
    \biggr\}.
\]
This is the transfer step: the product-law bound controls the event
before conditioning on the stimulus, and the native information controls
how much the channel can increase its probability.

It remains to obtain $\bar q$. The same dual penalty that defines the
hedge gives a product-law tail bound. Recall the expected perturbation
penalty from \eqref{eq:advset},
\[
    \Phi(C)
    := \frac{1}{\beta}\int\!\!\int P(s)P(a)f^\star(\beta C_s(a))\,da\,ds .
\]
For the regularizers below, $f^\star(y)\ge -1$. Since
$C_S(A)>u$ implies
$f^\star(\beta C_S(A))\ge f^\star(\beta u)$ on the positive tail, we have
\[
    \beta\Phi(C)
    \ge
    q f^\star(\beta u)+(1-q)(-1).
\]
Thus
\[
    q
    \le
    \frac{1+\beta\Phi(C)}
         {1+f^\star(\beta u)}.
\]
Following this method, the three regularizers give the following
bounds.

\paragraph{KL.}
For KL,
\[
    f(x)=x\log x-x+1,
    \qquad
    f^\star(y)=e^y-1 .
\]
Hence
\[
    \bar q_{\mathrm{KL}}(u)
    =
    \min\left\{
        1,\,
        \bigl(1+\beta\Phi(C)\bigr)e^{-\beta u}
    \right\}.
\]
The binary divergence is
\[
    D_{\mathrm{KL}}(\mathrm{Bern}(p)\|\mathrm{Bern}(\bar q_{\mathrm{KL}}))
    =
    p\log\frac{p}{\bar q_{\mathrm{KL}}}
    +
    (1-p)\log\frac{1-p}{1-\bar q_{\mathrm{KL}}}.
\]
Therefore
\[
    P(C_S(A)>u)
    \le
    \sup\biggl\{
        p\in[\bar q_{\mathrm{KL}}(u),1]
        \biggm|
        p\log\frac{p}{\bar q_{\mathrm{KL}}(u)}
        + (1-p)\log\frac{1-p}{1-\bar q_{\mathrm{KL}}(u)}
        \le I_{\mathrm{KL}}(S;A)
    \biggr\}.
\]

\paragraph{Pearson $\chi^2$.}
For Pearson $\chi^2$,
\[
    f(x)=(x-1)^2,
    \qquad
    f^\star(y)=
    \begin{cases}
        y+\frac{y^2}{4}, & y\ge -2,\\
        -1, & y<-2 .
    \end{cases}
\]
On the positive tail,
\[
    f^\star(\beta u)
    =
    \beta u+\frac{\beta^2u^2}{4}.
\]
Hence
\[
    \bar q_{\chi^2}(u)
    =
    \min\left\{
        1,\,
        \frac{1+\beta\Phi(C)}
        {\bigl(1+\frac{\beta u}{2}\bigr)^2}
    \right\}.
\]
The binary divergence is
\[
    D_{\chi^2}(\mathrm{Bern}(p)\|\mathrm{Bern}(\bar q_{\chi^2}))
    =
    \frac{(p-\bar q_{\chi^2})^2}
         {\bar q_{\chi^2}(1-\bar q_{\chi^2})}.
\]
Solving the upper branch gives
\[
    P(C_S(A)>u)
    \le
    \min\left\{
        1,\,
        \bar q_{\chi^2}(u)
        +
        \sqrt{
            I_{\chi^2}(S;A)\,
            \bar q_{\chi^2}(u)
            \bigl(1-\bar q_{\chi^2}(u)\bigr)
        }
    \right\}.
\]

\paragraph{Squared Hellinger.}
For squared Hellinger,
\[
    f(x)=(\sqrt{x}-1)^2,
    \qquad
    f^\star(y)=\frac{y}{1-y},
    \qquad y<1 .
\]
Assume $0<\beta u<1$. Then
\[
    \bar q_{H^2}(u)
    =
    \min\left\{
        1,\,
        \bigl(1+\beta\Phi(C)\bigr)(1-\beta u)
    \right\}.
\]
The binary divergence is
\[
    D_{H^2}(\mathrm{Bern}(p)\|\mathrm{Bern}(\bar q_{H^2}))
    =
    (\sqrt p-\sqrt{\bar q_{H^2}})^2
    +
    (\sqrt{1-p}-\sqrt{1-\bar q_{H^2}})^2 .
\]
Solving the upper branch gives
\[
\begin{aligned}
    P(C_S(A)>u)
    &\le
    \min\Biggl\{
        1,\,
        \biggl(
            \sqrt{\bar q_{H^2}(u)}
            \biggl(1-\frac{I_{H^2}(S;A)}{2}\biggr) \\
    &\hspace{6em}
            +
            \sqrt{1-\bar q_{H^2}(u)}
            \sqrt{
                I_{H^2}(S;A)
                -
                \frac{I_{H^2}(S;A)^2}{4}
            }
        \biggr)^2
    \Biggr\}.
\end{aligned}
\]

\section{Additional $f$-divergence examples}
\label{app:additional-f-divergences}

\begin{figure}[!tb]
    \centering
    \includegraphics[width=\linewidth]{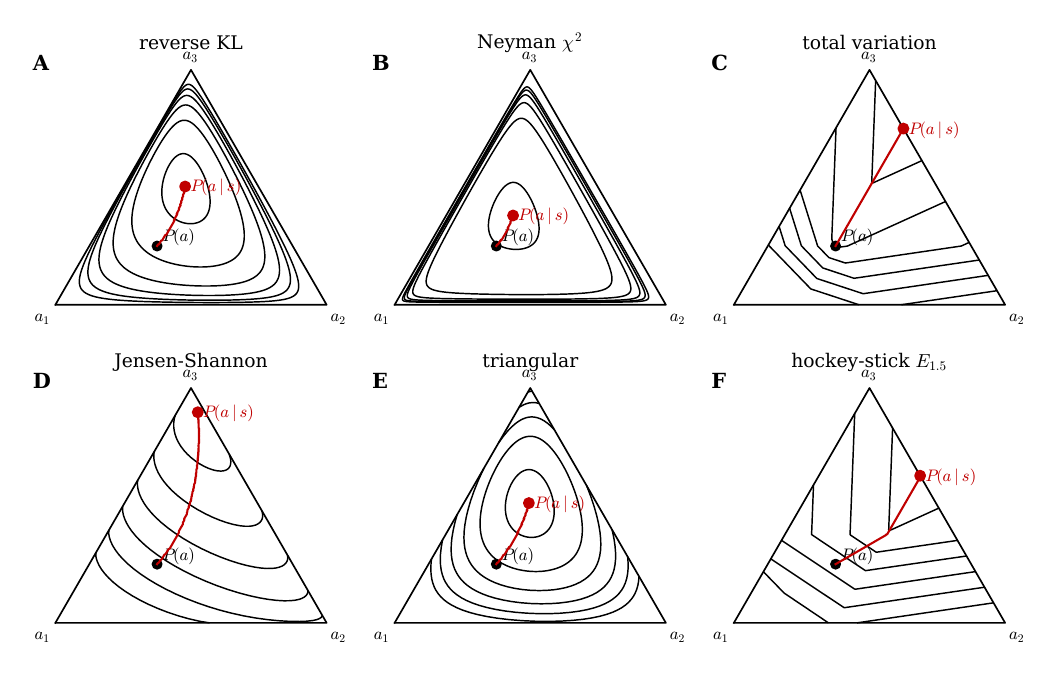}
    \caption{
    \textbf{Additional $f$-divergence geometries.}
    Each panel shows the bounded-rational objective
    on the three-action simplex for one of the regularizers in
    Table~\ref{tab:additional-f-divergences}. The black point is the
    marginal action distribution $P(a)$, the red curve is the operating
    path obtained by varying $\beta$, and the red point is the selected
    channel $P(a\mid s)$ at the displayed operating level. The panels
    illustrate that changing $f$ changes the native geometry of movement
    away from $P(a)$, and therefore changes the associated hedge.
    }
    \label{fig:additional-f-divergences}
\end{figure}

The examples in Table~\ref{tab:additional-f-divergences} and
Figure~\ref{fig:additional-f-divergences} were chosen to show how the
shape of $f$ changes the geometry of bounded-rational movement on the
simplex. All generators are written in canonical form, with $f(1)=0$
and, when differentiable at $1$, $f'(1)=0$. Throughout the table, $r = P(a|s)/P(a)$.

The main distinctions are determined by how $f(r)$ behaves near
$r=0$, near $r=1$, and as $r\to\infty$. Reverse KL and Neyman
$\chi^2$ place a large cost on suppressing actions that have positive
marginal probability, which bends the geometry away from the faces of
the simplex. Total variation and hockey-stick regularization introduce
kinks or flat regions, so the induced hedge is piecewise rather than
smooth. Jensen--Shannon and triangular discrimination grow slowly or
saturate in parts of the simplex, so large likelihood-ratio changes can
become relatively cheap compared with KL or Pearson $\chi^2$. These
differences are visible in the objective contours of
Figure~\ref{fig:additional-f-divergences}: the same loss and marginal
distribution produce different operating paths because each regularizer
changes which departures from $P(a)$ are cheap, costly, or ignored.

\begin{table}[!tb]
    \caption{Additional $f$-divergence regularizers.}
    \label{tab:additional-f-divergences}
    \centering\footnotesize
    \begin{tabular}{
        >{\raggedright\arraybackslash}m{0.12\linewidth}
        >{\centering\arraybackslash}m{0.34\linewidth}
        >{\centering\arraybackslash}m{0.24\linewidth}
        >{\raggedright\arraybackslash}m{0.20\linewidth}
    }
        \toprule
        Regularizer
        & $D_f(P(A|s)\|P(A))$
        & $f(r)$
        & Hedge behavior \\
        \midrule

        Reverse KL
        &
        \[
            \int P(a)\log\frac{P(a)}{P(a|s)}\,da
        \]
        &
        \[
            -\log r+r-1
        \]
        &
        Strong penalty as $r\downarrow 0$. \\[1.1em]

        Neyman $\chi^2$
        &
        \[
            \int
            \frac{\bigl(P(a|s)-P(a)\bigr)^2}{P(a|s)}
            \,da
        \]
        &
        \[
            \frac{(r-1)^2}{r}
        \]
        &
        Strong correction when an action is suppressed relative to $P(a)$. \\[1.1em]

        Total variation
        &
        \[
            \frac{1}{2}
            \int
            \bigl|P(a|s)-P(a)\bigr|\,da
        \]
        &
        \[
            \frac{1}{2}|r-1|
        \]
        &
        Nonsmooth hedge with piecewise-constant local correction. \\[1.1em]

        Jensen--Shannon
        &
        \[
            \begin{aligned}
            \int P(a)
            \Biggl[
                \frac{1}{2}\frac{P(a|s)}{P(a)}
                \log\frac{2P(a|s)}{P(a|s)+P(a)}
                \\
                +
                \frac{1}{2}
                \log\frac{2P(a)}{P(a|s)+P(a)}
            \Biggr]da
            \end{aligned}
        \]
        &
        \[
            \frac{1}{2}r\log\frac{2r}{1+r}
            +
            \frac{1}{2}\log\frac{2}{1+r}
        \]
        &
        Bounded information use; large likelihood-ratio changes saturate. \\[1.1em]

        Triangular
        &
        \[
            \frac{1}{2}
            \int
            \frac{\bigl(P(a|s)-P(a)\bigr)^2}
                 {P(a|s)+P(a)}
            \,da
        \]
        &
        \[
            \frac{(r-1)^2}{2(r+1)}
        \]
        &
        Smooth, moderate corrections; bounded as $r\downarrow 0$ and
        linear as $r\to\infty$. \\[1.1em]

        Hockey-stick $E_\gamma$
        &
        \[
            \int
            \bigl(P(a|s)-\gamma P(a)\bigr)_+\,da,
            \qquad \gamma>1
        \]
        &
        \[
            (r-\gamma)_+
        \]
        &
        Thresholded hedge; deviations below the likelihood-ratio threshold
        do not contribute. \\
        \bottomrule
    \end{tabular}
\end{table}

\section{Experimental Details}
\label{app:experiment}

Figure~\ref{fig:nominal-frontier} was generated from a fixed-design
regression problem. The design consisted of $n=100$ equally spaced
points $x_i\in[-1,1]$. For each Monte Carlo replicate, responses were
sampled independently as
\[
    y_i = 0.8\sin(\pi x_i)+0.25x_i+\epsilon_i,
    \qquad
    \epsilon_i\sim N(0,0.15^2).
\]
A training sample $s$ is the vector $(y_1,\ldots,y_n)$, and an action
$a$ is a prediction vector on the same design. The primitive loss was
mean squared error on this vector. The same loss was used for training,
loss scaling, and evaluation.

Two black-box learners were evaluated. The neural-network learner was a
one-hidden-layer ReLU network with $32$ hidden units and a scalar
$\tanh$ output, trained with Adam at learning rate $10^{-3}$. Its
operating control was training time, with $30$ geometrically spaced
checkpoints from $25$ to $3200$ gradient steps. The kernel learner was
kernel ridge regression with a Gaussian kernel of lengthscale $0.22$.
Its operating control was the ridge parameter, with $30$ geometrically
spaced values from $300$ to $0.3$.

To estimate the response law, prediction vectors were represented by a
shared finite action codebook. The codebook contained $200$ prediction
vectors obtained by $k$-means on pilot outputs from both learners,
across all operating settings and loss scales. Each prediction vector
was then assigned softly to the codebook by an exponential function of
squared prediction distance. This yielded an empirical conditional law
$P(A|S=s)$ at each operating setting; the marginal $P(A)$ was obtained
by averaging these conditional laws over Monte Carlo samples.

Certificates were estimated by the loss-scaling identity
\eqref{eq:estimate-certificate}. For each learner, operating setting,
and scale $t\in\{0,1/8,\ldots,1\}$, the learner was rerun with the
training loss multiplied by $t$, and the resulting action distribution
was evaluated using the original loss. The integral in
\eqref{eq:estimate-certificate} was computed by trapezoidal quadrature.
The native coordinate and operating parameter were then recovered along
each operating path using \eqref{eq:estimate-beta-info}. The Shannon
projection was computed from the same empirical response laws by
evaluating $I(S;A)/n$ directly.

The reported curves used $2000$ Monte Carlo training samples, $64$
pilot samples for the codebook, and $200$ bootstrap resamples. Bootstrap
intervals are the $5$--$95$ percentile intervals obtained by resampling
the Monte Carlo samples with replacement and recomputing losses,
certificates, native coordinates, and Shannon projections.

\end{document}